\documentclass[journal]{IEEEtran}

\usepackage[nospace,noadjust]{cite}
\usepackage[pdftex]{graphicx}
\usepackage{amsmath}
\usepackage{algorithmic}
\usepackage{array}
\usepackage[caption=false,font=footnotesize]{subfig}
\usepackage{dblfloatfix}
\usepackage[colorinlistoftodos]{todonotes}
\usepackage{textgreek}

\usepackage{tabularx}
\newcolumntype{g}{>{\hsize=1.2\hsize}X}
\newcolumntype{s}{>{\hsize=.8\hsize}X}

\begin{document}

\title{A Comprehensive Review of Shepherding as a Bio-inspired Swarm-Robotics Guidance Approach}

\author{Nathan~K.~Long,~\IEEEmembership{Student~Member,~IEEE,}
        ~Karl~Sammut,~\IEEEmembership{Senior~Member,~IEEE,}
        ~Daniel~Sgarioto,
        ~Matthew~Garratt,~\IEEEmembership{Senior~Member,~IEEE,}
        ~and~Hussein~Abbass,~\IEEEmembership{Fellow,~IEEE}%
\thanks{N. K. Long is with the School of Engineering and Information Technology, University of New South Wales, Canberra, ACT 2612, as well as the Maritime Division of the Australian Defence Science and Technology (DST) Group, VIC 3207, Australia. e-mail: nathan.k.long.91@gmail.com.}%
\thanks{K. Sammut is with the College of Science and Engineering at Flinders University, Tonsley, SA 5042, and with DST Group Maritime Division, SA 5111, Australia.}
\thanks{M. Garratt, and H. Abbass are with the School of Engineering and Information Technology, University of New South Wales, Canberra, ACT 2612, Australia.}%
\thanks{D. Sgarioto is in the Maritime Division of DST Group, Port Melbourne, VIC 3207, Australia.}%
}


\maketitle

\textit{Accepted Article: Copyright 2020 IEEE. Personal use of this material is permitted. Permission from IEEE must be obtained for all other uses, in any current or future media, including reprinting/republishing this material for advertising or promotional purposes, creating new collective works, for resale or redistribution to servers or lists, or reuse of any copyrighted component of this work in other works.} \newline

\begin{abstract}
The simultaneous control of multiple coordinated robotic agents represents an elaborate problem. If solved, however, the interaction between the agents can lead to solutions to sophisticated problems. The concept of swarming, inspired by nature, can be described as the emergence of complex system-level behaviors from the interactions of relatively elementary agents. Due to the effectiveness of solutions found in nature, bio-inspired swarming-based control techniques are receiving a lot of attention in robotics. One method, known as swarm shepherding, is founded on the sheep herding behavior exhibited by sheepdogs, where a swarm of relatively simple agents are governed by a shepherd (or shepherds) which is responsible for high-level guidance and planning. Many studies have been conducted on shepherding as a control technique, ranging from the replication of sheep herding via simulation, to the control of uninhabited vehicles and robots for a variety of applications. A comprehensive review of the literature on swarm shepherding is presented in order to reveal the advantages and potential of the approach to be applied to a plethora of robotic systems in the future.
\end{abstract}


\IEEEpeerreviewmaketitle

\section{Introduction}

\IEEEPARstart{I}{nspired} by animal swarms in nature, swarm robotic systems have the potential to offer simple, dynamic solutions to complex problems. Swarms are capable of emulating and augmenting multi-agent dynamic systems to complete intricate, objective-based behaviors with relatively simple local agents~\cite{Zoghby2013}. This manifestation of intelligent behavior without external regulation is described as an emergent property~\cite{Harvey2018}. Swarm robotic systems permit decentralization of the control system, where each agent is responsible for its own processing and functionality, allowing simpler agents to be designed rather than an all-encompassing complex single agent~\cite{Parker1998}. Therefore, decentralization makes robotic systems easier to build and program, as well as introducing robustness, flexibility, and scalability into the multi-robot system~\cite{Harvey2018}.

Controlling multiple coordinated robotic agents presents a complex problem which can be categorized into high-level path planning and low-level single agent dynamics. While various swarm control techniques have been developed, including using neighbourhood topologies~\cite{Gazi2011}, artificial potential functions~\cite{Gazi2005},~\cite{Gazi2012}, and different forms of adaptive control~\cite{Lim2009},~\cite{Gazi2015}; one method, termed shepherding, has emerged which divides the two control functions into separate entities. Vo, Harrison and Lien~\cite{Vo2009} defined the motion planning of a group of robots using an external entity as the \textit{group control problem}, describing it as underactuated and dynamic, which elicits multi-agent cooperation. 

Shepherding, as a swarm control technique, was inspired by sheepdogs and sheep, where the shepherding problem can be defined as the guidance of a swarm of agents from an initial location to a target location~\cite{Lien2004}. Essentially, the shepherd is responsible for the high-level path planning of the swarm and task allocation, while the swarm members themselves function solely on single agent dynamics.

It should be noted that while previous literature has referred to the group of agents guided by the shepherd as a \textit{flock} or \textit{herd}, this review uses the term \textit{swarm} for two reasons. First, the modeling of a flock is mostly done using swarm intelligence principles and rules. Second, the term swarm generalizes the common characteristics exhibited by a flock, a herd, a school of fish, and other biologically or behaviorally inspired terminologies used in the literature.

Many applications for multi-robotic teams controlled via shepherding have been discussed, covering a wide selection of domains. Of interest, these include the use of robotic agents to herd sheep~\cite{Evered2014},~\cite{Bat-Erdene2017}; crowd control~\cite{Lien2009}; oil spill cleanups~\cite{Masehian2015}; protecting aircraft from bird strikes~\cite{Gade2015,Gade2016,Paranjape2018}; disaster relief and rescue operations~\cite{Shedied2013}; fishing~\cite{Masehian2015}; managing wildlife distribution and protection~\cite{Pierson2015}; and the handling of micro-organisms~\cite{Ozdemir2017}. Furthermore, shepherding has been examined for security and military procedures, such as uninhabited vehicle (UxV) maneuvering in combat~\cite{Chaimowicz2007},~\cite{Razali2012}, searching terrains for intruders~\cite{Shedied2013}, and mine collection and distribution~\cite{Masehian2015}.

This review paper covers the majority of shepherding literature, as a swarm control method, to date, with the objective of using previous research to fortify the notion that shepherding can act as simple, yet effective, means for controlling a wide variety of swarm robotic systems in the future. 

The survey is organized as follows. Section \ref{sec:trials} gives a brief overview of real techniques and assessments of sheepdog shepherding methods. This is used to build a base knowledge of the biological origins of technical shepherding. Section \ref{sec:found} outlines the early studies on shepherding as a swarm control technique which laid the foundations for the research that followed. Section \ref{sec:class} details the classic shepherd herding sheep inspired studies. Section \ref{sec:control} describes the specific control techniques used for shepherding in more detail. Section \ref{sec:alt} illustrates alternative behaviors used in shepherding, while Section \ref{sec:rob} highlights the shepherding studies into actually using UxVs and robots, including those which involve animal shepherding. Section \ref{sec:hum} describes the research into human-shepherd interactions, and Section \ref{sec:ML} showcases how machine learning and computational intelligence techniques have been applied to shepherding research. Finally, Section \ref{sec:chal} articulates some of the research challenges facing shepherding swarm control.

\section{Sheepdog Trials} \label{sec:trials}

The International Sheep Dog Society\textquoteright s Rules for Trials~\cite{IntSheepdogSoc2018} are designed to measure the ability of a sheepdog to perform the basic skills required for shepherding. From a modeling, machine learning and artificial intelligence (AI) perspective, these rules offer four main advantages. The first advantage is the ground work that has evolved over many centuries in this domain, which decomposes the shepherding problem into the basic sub-skills required for shepherding. The second advantage is in the possible use of the scenarios used in these competitions as test benchmarks for shepherding. The third advantage lies in the nature of these competitions, where the sheepdog needs to interact with a handler (a human) to shepherd the sheep. Replacing a biological sheepdog with a smart autonomous system that acts as a smart artificial sheepdog would mean that the scenarios used in the competition could also be used for testing the efficacy of the human-AI interaction. Last, but not least, the rules include a point-based system to assess the dogs which could be useful for the design of reward functions for an AI to shepherd.

The rules list and test for the basic skills required by a sheepdog to shepherd. These skills are summarized below:

\begin{itemize}
\item \textbf{Outrunning}: The handler stands at the post. As soon as the dog receives an order, the dog is expected to perform a pear shaped run that gets wider as the dog approaches the flock. At the end of the outrun, the dog is expected to be positioned behind the sheep at the \textit{point of balance}, with sufficient distance so as not to disturb the sheep, while equally in a close enough proximity to gain control.

\item \textbf{Lifting}: Lifting commences when the dog starts to influence the sheep, getting them to move in an orderly manner.

\item \textbf{Fetching}: The dog is expected to drive the sheep in the direction of the handler, while ensuring that the whole flock passes through the fetch gates. If any sheep does not pass through the gate, it is considered a failed trial.

\item \textbf{Driving}: Once fetching is complete and the flock are driven around the handler, the sheepdog is expected to drive the flock away from the handler towards the first drive gates.

\item \textbf{Shedding}: When the flock passes through the second drive gates, the sheepdog is expected to turn the sheep towards the shedding ring, where the handler and the sheepdog sort and separate the flock into a specified number of sheep. 

\item \textbf{Penning}: The flock is driven into a small enclosure, where the handler is responsible for closing the gate once all sheep are inside. 

\item \textbf{Singling}: Similar to, but more challenging than, shedding, where a specific single sheep needs to be separated from the flock and driven away.

\item \textbf{Gathering/Collecting}: The collecting behaviour is when the sheepdog brings the flock to the handler.
\end{itemize}

Another list of sheepdog commands were defined by Bennett and Trafankowski~\cite{Bennett2012} below, where each command represents an expected behaviour from the sheepdog.

\begin{itemize}

\item \textbf{Stay}: Dog halts and remains still.

\item \textbf{Lie Down}: Dog lies down.

\item \textbf{Easy, Steady, or Take Time}: The dog moves slowly.

\item \textbf{Walk-Up}: The dog approaches the nearest sheep and stops at a distance sufficiently large enough to not exert any force on the sheep to flee.

\item \textbf{Look Back}: The dog leaves the current group of sheep and searches for others. 

\item \textbf{Get Back or Get Out}: The dog moves back away from the flock.

\item \textbf{That will do}: The dog returns to the handler.

\item \textbf{Come-bye}: The dog circles the flock in a clock-wise manner.

\item \textbf{Away to Me}: The dog circles the flock in a counter-clock-wise manner.
\end{itemize}
 
The above commands could be seen as the basic skills required by a sheepdog to herd. The higher order planning is conducted by the handler, who then decides which command to issue to the sheepdog. As such, the sheepdog could be seen to exhibit a basic level of autonomy once given a particular command, while the handler is the smarter entity that decides which command to issue and when. 

Keil~\cite{Keil2015} saw the handler-dog relationship as an inter-species distributed cognitive system, whereby the handler extends the limited cognitive skills of a sheepdog by scaffolding the actions needed to complete the task. Having focused on the cognitive dimension of the relationship, Keil did not explore the physical augmentation dimension, where the sheepdog extends the handler with its physical abilities. This tight coupling between the handler and sheepdog allows the human-animal augmented system to carry out tasks that neither of the two systems in isolation could do alone. Keil saw the role of the handler to offer higher order cognition and executive control skills that a sheepdog does not possess. The human-sheepdog relationship relies on a coordinated pattern of movements relative to sheep and commands. The latter mediates the sheepdog-sheep relationship to moderate the pressure exerted on the sheep. The handler needs to estimate the point and level of pressure that need to be exerted on the sheep. The commands are seen by Keil as the mean to connect the physical skills of the sheepdog with the cognitive skills of the human.

Sheep rely on a multitude of behavioral, cognitive and physical traits to protect themselves. They rarely walk in a straight line, are able to direct their ears to the direction of the sound and to smell predators, have a high tolerance to pain, and flock in numbers. Sheep are known to have a hearing frequency of 100Hz to 30KHz. They possess a panoramic field of 320-340 degrees, but with low perception of depth, only have a binocular field of 20-60 degrees. Moreover, sheep have relatively good memory with an ability to recollect up to 50 unique human or sheep faces for up to two years~\cite{S1012019}. Werner and Dyer~\cite{Werner1993} published a concise summary on the advantages of herding. They discussed the many-eyes theory, whereby the herd uses all eyes in the herd to look and sense for a threat. We can extrapolate this concept to a sensor array of all eyes, noses and ears. A herd has several advantages when compared with individuals, such as being able to find food more easily and creating the social structure to access mates. From a safety perspective, herding offers a stronger line of defence and protection, and can intimidate an intruder.

In comparison, sheepdogs come in different breeds with various skills and attributes. A sheepdog is considered to be clever if it is able to understand new commands in less than five repetitions with the ability to obey first command 95\% of the time~\cite{coren2006intelligence}. A range of dog breeds are, thus, considered adept at becoming sheepdogs, including Border Collies, Poodles, German Shepherd Dogs, Golden Retrievers, and Doberman Pinschers. Trained sheepdogs can sell for as much as AU\$25,000~\cite{DHT2019}. While a sheepdog may be more than just a work tool, this figure highlights the value of such multi-agent intelligent problem solving systems.
 
\section{Shepherding Foundations} \label{sec:found}

In the late 1990s, a team of researchers launched the Robot Sheepdog Project~\cite{Vaughan1997}. The Robot Sheepdog Project aimed to create a robot capable of controlling a swarm of animals, initially envisaged as sheep, but to reduce the complexity of experimentation inherent in dealing with relatively large animals, ducks were chosen. A custom-made robot used a mounted-camera to track the duck swarm and control its movement. The robot approached the ducks, causing them to group together, then repulsed them in the direction of a goal location. The Robot Sheepdog Project went on to produce three more studies~\cite{Vaughan1998,Sumpter1998,Vaughan2000}, on their research. 

Sumpter et al.~\cite{Sumpter1998} developed a technique using machine vision, via image sequencing, to categorize the behaviors of ducks being controlled by the robot and, subsequently, to predict future animal motion. Vaughan et al.~\cite{Vaughan1998} then created an algorithm which replicated the flocking behaviour of ducks, and used it to design a control system for the shepherding robot. The algorithm successfully guided a swarm of real ducks to a goal location.

In contemporaneous work, Schultz, Grefenstette and Adams~\cite{Schultz1999} used genetic algorithms to create decision rules to replicate shepherding behavior, where rule sets were developed through simulation. After 250 generations, the authors experimented with robots, where one robot was used to control another. The results showed that the robotic shepherd was able to guide the swarm agent to a goal destination in 67$\%$ of the trials. 

Diverging from classic sheepdog herding sheep studies, Funge, Tu and Terzopoulos~\cite{Funge1999} created a simulation in which a \textit{T-Rex} herded a swarm of \textit{Raptors} from its territory to another. The purpose of the research was to develop a cognitive model to enhance behavior-based models to govern characters for computer animations. The shepherding T-rex was able to use information on its actions within the environment, and their effects, to enact objective-based behaviors with positive results.  

These studies~(\cite{Vaughan1997,Vaughan1998,Sumpter1998,Vaughan2000,Schultz1999,Funge1999}), were some of the earliest shepherding research produced, and are quite representative of the studies to follow. Simulation and robotic experimentation have been conducted to try to recreate the bio-originated shepherding behavior seen in sheepdogs, while AI techniques, such as machine learning and genetic algorithms, have been used to augment the shepherding task. Moreover, shepherding has been used as a swarm control technique for a variety of other swarm-based applications.

\section{Classic Shepherding} \label{sec:class}

The most common type of shepherding research has been the replication of the sheepdog herding sheep behaviors via computer simulation. Artificial sheepdogs will be referred to as shepherds, artificial sheep agents as a
swarm or swarm agents, and the replication of the behavior of sheepdogs guiding sheep to an objective location as herding for the rest of the text. Kachroo et al.~\cite{Kachroo2001} created two dynamic programming algorithms used to guide a swarm to an objective location using a shepherd (represented as objects), where the swarm only reacted to the shepherd. One of the algorithms was shown to perform better than the other, while a complexity analysis was performed for each. A description of the software package developed for the study was also presented. A follow up study by Kacharoo, Shedied and Vanlandingham~\cite{Kachroo2002} then introduced stochastic motion into the swarm agents, developing three dynamic programming algorithms to solve the shepherding problem. Using an optimal cost value function, it was shown that two of the algorithms performed approximately equally, while the third was significantly more time consuming. 

A simple algorithm presented by Miki and Nakamura~\cite{Miki2006} was used to simulate a herding task. The authors describe four rules for the behavior of the swarm: \textit{cohesion} to the nearest neighbor, \textit{separation} to avoid collisions with obstacles or other swarm agents, \textit{escape} away from the shepherd(s), and \textit{random action} which creates random stochastic movements. Likewise, the shepherd was governed by four rules: \textit{guidance} of the flock in an objective direction, \textit{flock making} to push stray sheep back to the group, \textit{keeping} a certain distance from the flock to avoid splitting it, and \textit{cooperation} to avoid shepherds overlapping. The movement of the shepherd was dictated by its coordinates relative to the objective direction of the sheep swarm, and sheep center. The results showed the algorithms effectively replicated herding behaviors, and were compared for one and two shepherds which consistently showed the two shepherds outperforming only one at guiding the swarm. 

Str\"{o}mbom et al.~\cite{Strombom2014} designed a simple heuristic for mimicking sheep behaviors, using only a single shepherd, which closely matched that of~\cite{Miki2006}, where the two-dimensional (2D) simulations were comprised of similar behavioral rules. Differences included swarm attraction to the centre of mass (CM) of their local neighborhood (instead of just single neighbor), and the solo-shepherd not requiring cooperation with another shepherd, as well as the terms guidance and flock making replaced with \textit{driving} and \textit{collecting}. Further, the shepherd\textquoteright s motion was less strategically planned in~\cite{Strombom2014} with respect to avoiding unwanted influence on the swarm when moving to a driving or collecting position. This likely resulting in additional sheep agent and sheepdog motion, thus, increasing energy expenditure. Figure \ref{fig_strombom} gives an example of a sheepdog driving sheep towards an objective location using Str\"{o}mbom et al.\textquoteright s heuristic (discussed further in Section \ref{sec:control}). 

In 2018, Hoshi et al. conducted two studies~\cite{Hoshi2018},~\cite{Hoshi20182}, which elaborated on Str\"{o}mbom et al.\textquoteright s work~\cite{Strombom2014}. The first~\cite{Hoshi2018} tested the effects of varying the step size per time step of the shepherd and swarm agents on the shepherding task. The task was performed much better when the shepherd was able to move faster than a swarm which moved at the same speed. Further, the shepherd was generally able to herd swarm agents which moved at variable speeds, except when the discrepancy between the average and slowest agent was great. In the second study~\cite{Hoshi20182}, the shepherd and swarm agent\textquoteright s motion and influential force vectors were extended to the third dimension. The shepherd was found to be able to guide the swarm moving at a greater range of speeds in the three-dimensional (3D) simulation than in the 2D simulation, while performing similarly for a swarm moving at variable speeds. Hoshi et al.\textquoteright s studies increase the fidelity of Str\"{o}mbom et al.\textquoteright s~\cite{Strombom2014} simulation by introducing irregularities in the swarm agent\textquoteright s motion, better representing the motion of real animals or robotic agents (due to aggregated errors and assumption-based inaccuracies in their code)~\cite{DuToit2012}. The authors\textquoteright \ also showed the applicability of shepherding in the third dimension, though not particularly for the case of shepherding land-based livestock; however, potentially for water or air-based animals. 

Another algorithm, developed by Lee and Kim~\cite{Lee2017}, implemented an alternative set of swarm and shepherd behaviors to solve a shepherding problem. If swarm agents are within a specified distance from each other, then they try to minimize the distance between themselves. The agents then try to move to the center of their swarm neighborhood, while attempting to match their neighbors\textquoteright \ velocities. Finally, the swarm agents avoid collisions with both obstacles and the shepherds. The shepherds collect scattered agents into a single swarm, then steer them towards a goal destination. Lee and Kim, further, incorporated \textit{patrolling} (keeping the swarm at a fixed position) and \textit{covering} (navigating the swarm to multiple goals) shepherd behaviors into their simulations. An extensive methodology and discussion was included, and their algorithm was shown to be effective at autonomously controlling the swarm. 

Tsunoda, Sueoka and Osuka~\cite{Tsunoda2017} used similar swarm agent behaviors to~\cite{Strombom2014}; however, with an introduced angle of error for the shepherd to swarm agent repulsion, where the sheep would not move directly away from the shepherd, rather, in a direction with an offset angle. The angles tested were based on a normal distribution of varying standard deviation (SD). Once the SD was increased to $8^\circ$, the shepherd struggled to effectively guide the swarm agents to the goal location. Further results showcased the necessity of correct parameterization, and the increased performance of the shepherd as its speed increased. 

Of note, in many studies it was found that a zig-zag motion, used by the shepherd to drive and collect the swarm agents, appeared to be an efficient guidance method as the behaviour emerged in~\cite{Miki2006,Strombom2014,Hoshi2018}, while also being enforced in~\cite{Lien2004},~\cite{Bennett2012},~\cite{Fujioka2016},~and~\cite{Fujioka2018}.

\section{Shepherding Control Methods} \label{sec:control}

In an influential study by Reynolds~\cite{Reynolds1987} in 1987, one of the first swarm control techniques was developed, which was used to replicate the flocking behavior of birds in computer graphics. Reynolds represented each member of the flock as a \textit{boid}, where the boid was represented as an object which can sense both its location and orientation. Reynolds used a series of acceleration (force) vectors to simulate three responses of the bird swarm: \textit{collision avoidance} to ensure each bird did not collide with obstacles, including other birds, \textit{velocity matching} so that each bird tried to align with the direction and speed of other birds in its neighborhood, and \textit{flock centering} to maintain a coherent swarm by clustering together. 

\begin{figure}[!t]
\centering
\includegraphics[scale=0.32]{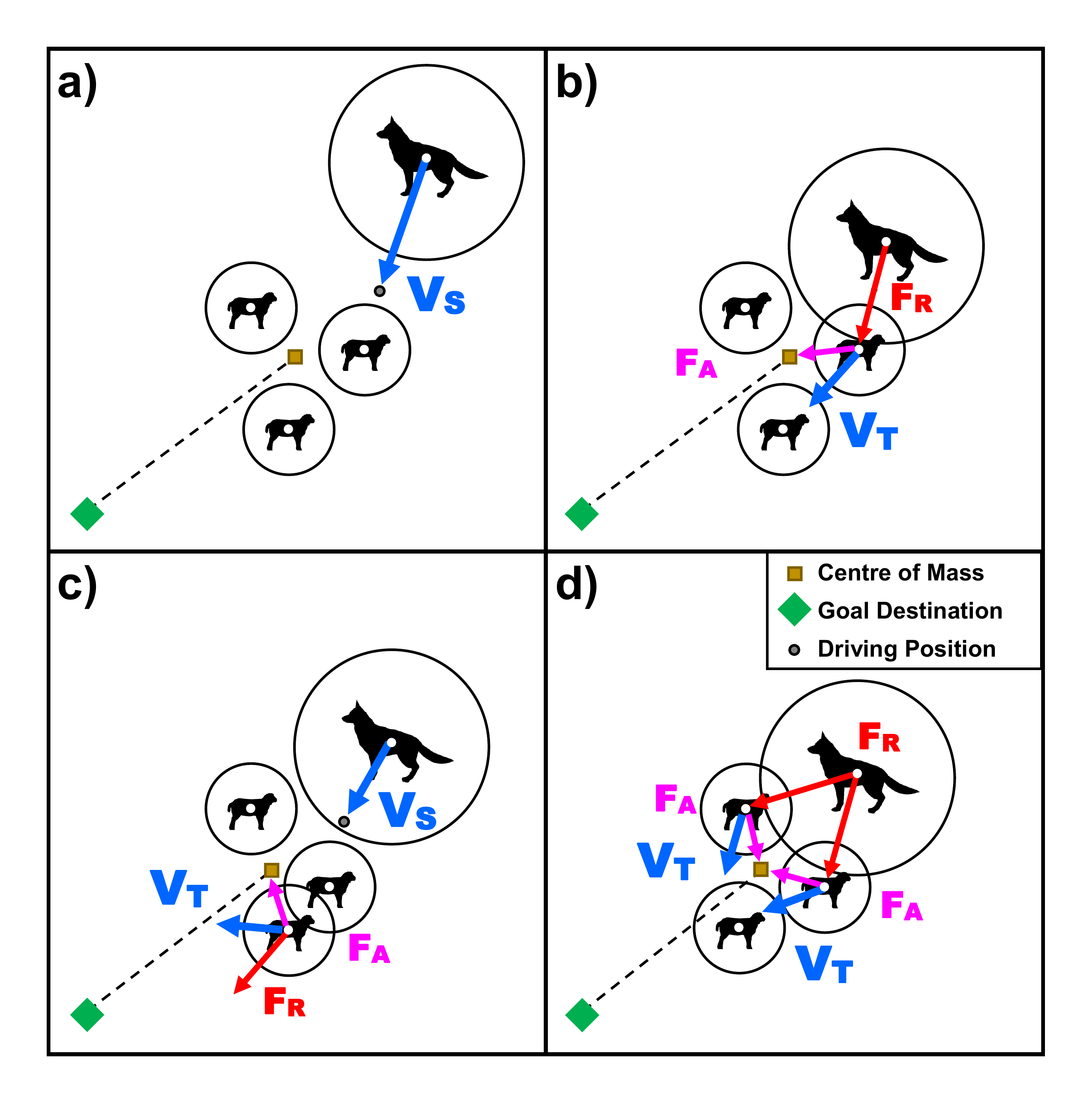}
\caption{Schematic outlining Str\"{o}mbom et al.\textquoteright s~\cite{Strombom2014} shepherding heuristic, where the circles represents radii of influence, and arrows represent vectors. a) The shepherd approaches the driving position. b) The shepherd repels a sheep, which is also attracted to its CM. c) The sheep repels another sheep, which is also attracted to their CM, while the shepherd moves to the new driving position. d) The shepherd repels two sheep, which are also both attracted to their CM.}
\label{fig_strombom}
\end{figure}

The idea of using force vectors to control a swarm of agents has since become fundamental for shepherding control, particularly the swarm\textquoteright s repulsion from a shepherd (or shepherds). For example, Str\"{o}mbom et al.~\cite{Strombom2014} applied Reynolds\textquoteright \ boid and the force vector model~\cite{Reynolds1987} to their shepherding heuristic to enforce the sheep agent and shepherd behaviors, as shown in Figure \ref{fig_strombom}. The schematic illustrates a shepherd guiding a swarm of sheep towards a goal destination, where $V_S$ represents the shepherd\textquoteright s velocity vector, $F_R$ represents the repulsion force of the shepherd, $F_A$ represents the attraction force to the sheep\textquoteright s CM, and $V_T$ represents the total velocity vector for a sheep.

Twelve years later, another paper was published by Reynolds~\cite{Reynolds1999}, which described several \textit{steering} behaviors used to control autonomous agents, modeled as simple point-mass vehicles. One of these behaviors, \textit{evasion} was analogous to shepherding. Evasion involved an agent aligning its velocity vector in the opposite direction of the predicted future location of another agent. Other behaviors described by Reynolds, executed using force vectors, include \textit{seek}, \textit{flee}, \textit{pursuit}, \textit{arrival}, \textit{obstacle avoidance}, \textit{wander}, \textit{path following}, \textit{containment}, \textit{flow field following}, \textit{separation}, \textit{cohesion}, \textit{alignment}, \textit{flocking}, and \textit{leader following}. Reynolds provided a good description of the application of each behavior; however, gives no physical or simulated results to validate his descriptions. 

Bayazit, Lien and Amato~\cite{Bayazit2002} developed a rule-based roadmap control system to produce \textit{homing}, \textit{exploring}, and \textit{shepherding} swarm behaviors. Each roadmap was comprised of a series of nodes connected by edges. For the homing and exploring behaviors, the swarm agents communicate the potential of their local path by weighting the edges, then the agents follow the most preferential edge at nodes with multiple paths. However, when shepherding, the shepherd would calculate a path towards an objective location, then allocate nodes along the path as sub-goals. If the swarm was too separated, then the shepherd would move to a node to realign a swarm sub-group with the rest of the swarm. Simulated results for shepherding, using 30 swarm agents and a single shepherd, showed faster convergence to a goal when compared with an $A^*$ path planning technique. Bayazit, Lien and Amato~\cite{Bayazit2004} published a follow-up study on their roadmap design which did not include shepherding; however, included a technique for narrow passageway navigation. Their control system was then applied to two shepherding studies by Lien et al.~\cite{Lien2004},~\cite{Lien2005}. Vo, Harrison and Lien~\cite{Vo2009} then tried to extend the roadmap-based control system for shepherding through a variety of methods; however, while the elaborated path planning broadened the range of environments where their control technique was effective, it was unable to perform well generically. 

Following their previous research (\cite{Vo2009},~\cite{Lien2004},~\cite{Lien2009},~\cite{Lien2005}), Harrison, Vo and Lien~\cite{Harrison2010} produced a new control method for shepherding a swarm, in which the shepherd models the swarm as an abstract, deformable shape. Using this abstraction, the shepherd continually updates the current swarm shape by calculating a new desired distribution, then moving to steering points behind the swarm (with respect to the goal location) to achieve it. The results showed that the method outperformed the authors\textquoteright \ previous shepherding control techniques.

A unique shepherding control system was created by Razali, Meng and Yang~\cite{Razali2010}, inspired by the theory of immune networks, which permits a distributed control system, and is capable of adapting dynamically to an environment. The memory-based immune system acted as an analogy for the shepherding problem, where obstacles are represented as antigens, robots as B-cells, and actions as antibodies. The results of the study, while not discussed in significant detail, did indicate that the method was ineffective at shepherding robotic swarm agents. A follow-up study by Razali, Meng and Yang~\cite{Razali2012}, however, provided a more in-depth discussion of their method. The simulated shepherds cooperated with each other via the immune system-based approach, while using a \textit{vector field histogram} algorithm to avoid obstacle collision and instigate objective-based behaviors. The simulated swarm were reported as exhibiting \textit{avoidance} and \textit{flocking} behaviors. The authors\textquoteright \ second study enforced a refined alignment of the shepherds behind the swarm agents for improved control. While enhanced results are presented, the technique was only applied to swarms of two to four agents, and showed that the agents could not all be shepherded and contained within a target region.

Bacon and Olgac~\cite{Bacon2011}, in contrast to previously mentioned studies, developed a control system which used multiple \textit{pursuers} (shepherds) to guide a single \textit{evader} (swarm agent) along a predefined trajectory. This was implemented by distributing the pursuers along an arc, centered on the evader, using sliding mode controllers. The evader is guided, and pursuers spaced, by a repulsion force. Pierson and Schwager~\cite{Pierson2015} used a similar multi-shepherd control strategy, where the shepherds guided a swarm agent by distributing themselves along an arc centered on the agent, via a single continuous control law which negated the need for behavioral switching. Further, the authors elaborated their design to include multiple swarm agents, with the swarm\textquoteright s mean position at the origin of the arc. Including more shepherds than swarm agents, however, increases the complexity of the shepherding task, and reduces the decentralization of the control system. Pierson and Schwager~\cite{Pierson2018} then elaborated upon their previous study by modifying their control system for 2D environments with obstacles, then for a 3D environment (creating a 3D spiral arc around the swarm). In order to avoid the introduced obstacles in the 2D environment, a series of sub-goal locations were introduced such that the shepherd would guide the swarm along a path which navigated around the obstacles. The authors 3D simulations showed seven shepherds effectively guiding three swarm agents to an objective location, as well as six shepherds guiding an undisclosed swarm size around obstacles to a goal location, in a similar manner to the 2D simulations. While their publication shows the 2D simulations working for fewer shepherds than swarm agents, few details are provided on the range of agents tested.  

An algorithm, developed by Bennett and Trafankowski~\cite{Bennett2012}, took inspiration from commands used by humans to control sheepdogs (described in Section \ref{sec:trials}), as well as previous swarm shepherding techniques. The swarm agents were repulsed by the shepherd, other swarm agents, and obstacles. The swarm attraction was varied between being to their closest neighbour, the centroid of the swarm, to a random swarm member, and to all other swarm agents. The shepherd guided the swarm towards a goal by circling it in the clockwise direction if the swarm centroid was left of the goal (relative to itself), or anti-clockwise if the centroid was to the right of the goal. It then followed the circular arc until it was positioned a certain angle from the furthest left or right swarm agent from the centroid (with respect to the goal). Different parameter values were tested, then the algorithm was compared to the performance of~\cite{Lien2004},~\cite{Vaughan2000},~and~\cite{Miki2006}. Vaughan et al.\textquoteright s study~\cite{Vaughan2000} was shown to perform very poorly in an open environment, while Lien et al.\textquoteright s~\cite{Lien2004} and Miki and Nakamura\textquoteright s~\cite{Miki2006} algorithms were generally effective; however, performance reduced when the swarm agents were weakly attracted. The author\textquoteright s algorithm gave good results; however, was very sensitive to the angle that caused it to switch directions. The comparison was done by recreating the algorithms from~\cite{Lien2004},~\cite{Vaughan2000},~and~\cite{Miki2006}, so results may have varied from their original codes. 

A study by Shedied~\cite{Shedied2013} optimized the trajectory of a shepherd guiding a robotic agent to an objective location, where the shepherd and agent were both considered to be wheeled mobile robots (WMRs). The distance between the WMRs was fixed, then the relative angle between them at each time step was optimized to obtain a complete trajectory. Once the optimal trajectory was found, the author introduced non-holonomic constraints, specifically the robot\textquoteright s angular and linear velocity constraints, such that their optimal values at each time step could be determined so that the WMRs followed the proposed trajectory. While the trajectory optimization was practical for a single agent, it did not extend to the control of an entire swarm. Thus, it is difficult to analyse its applicability to multi-robotic systems.

\begin{figure}[!t]
\centering
\includegraphics[scale=0.4]{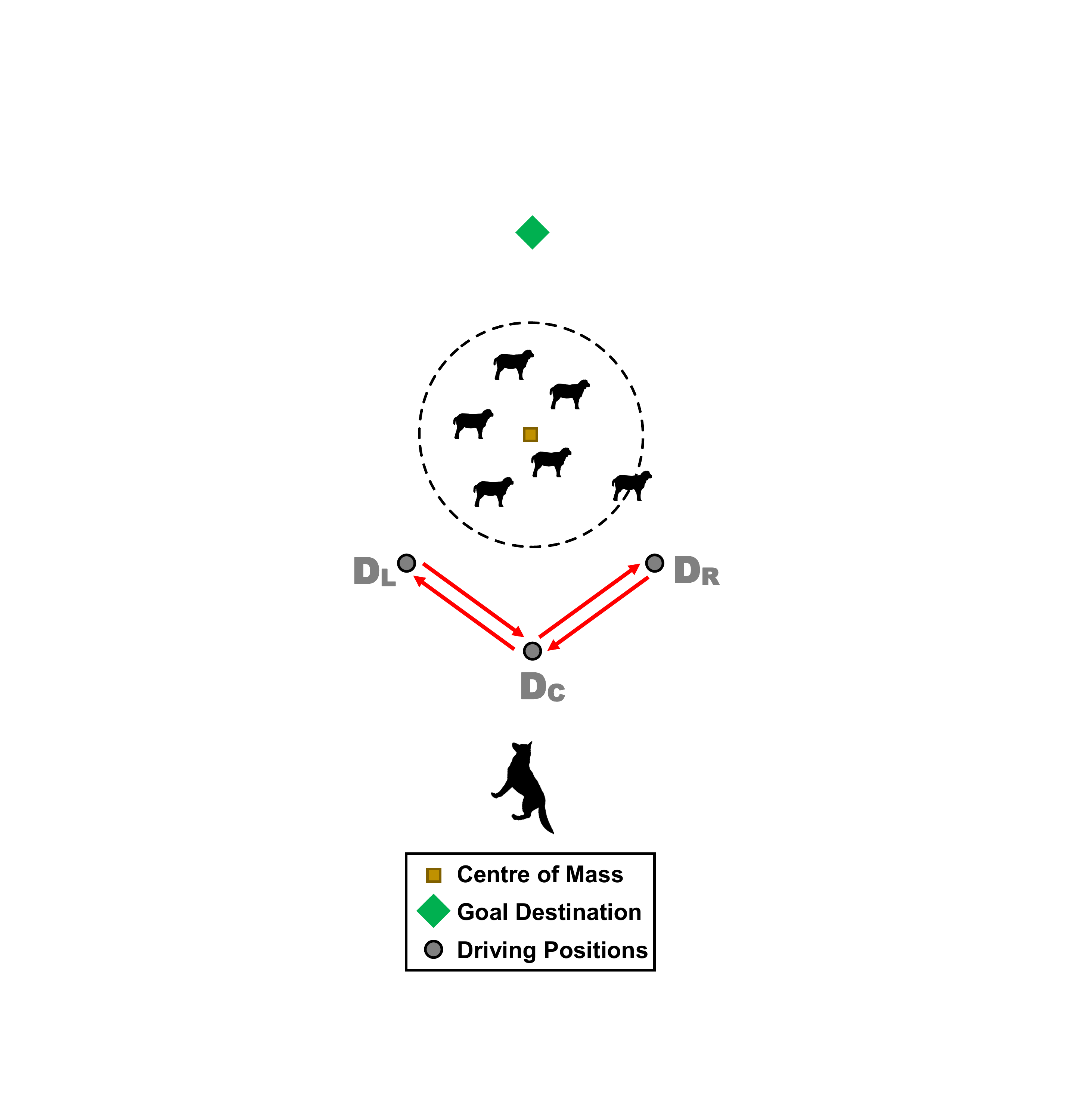}
\caption{Schematic outlining Fujioka and Hayashi\textquoteright s~\cite{Fujioka2016} V-formation control method, where $D_L$, $D_C$, and $D_R$ represent the left, right, and center driving positions of the formation. The dashed circle indicates the radius from the CM of the furthest swarm agent, outlining the extra distance of the driving positions.}
\label{fig_fujioka}
\end{figure}

Fujioka and Hayashi~\cite{Fujioka2016} introduced an alternative shepherding control behavior called \textit{V-formation control}, where the shepherd cycles between three positions along an arc extending out from a swarm\textquoteright s CM, which were centered behind, and rotated slightly to the left and right ($<90^\circ$ from the center position), creating a V-shaped trajectory, shown in Figure \ref{fig_fujioka}. The authors then combined Str\"{o}mbom et al.\textquoteright s~\cite{Strombom2014} approach with their own, such that the collecting behavior of~\cite{Strombom2014} was replaced with the V-formation control. The results showed that Fujioka and Hayashi\textquoteright s methods had greater shepherding task success rates; however, generally took longer to achieve them, than Str\"{o}mbom et al.\textquoteright s. Fujioka~\cite{Fujioka2018} then elaborated on the previous V-formation control method to include two shepherds, where six points along the arc were used, with one shepherd performing the V-formation to the left of the center behind the CM with respect to the goal, and the other to the right of the center (where the angle between each position is $<45^\circ$). Moreover, Fujioka tested a method they called \textit{shift V-formation control}, which sees the original single-shepherd, three-point V-formation shift about the CM depending on where the furthest swarm agent is positioned in order to collect it. Results indicated that two shepherds were able to better control the swarm and to guide them to the objective location in fewer time steps than one. The original V-formation control was also shown to be more effective than the shift V-formation control technique. 

While most shepherding control approaches assume that the shepherd(s) have global knowledge of the swarm agent positions, Tsunado et al.~\cite{Tsunoda2018} tested a more sheepdog-realistic scenario in which the shepherd only had swarm position information collected via a simulated local camera. Three different shepherding control strategies were used. The first was the \textit{center-of-group targeting controller} used in~\cite{Tsunoda2017}, the second was the \textit{online target switching controller}, the same as used in~\cite{Strombom2014}, and the final method was the \textit{furthest agent targeting}, where the shepherd constantly aims to repel the furthest swarm agent from the goal location towards it. The results for the local camera-based shepherding showed that the shepherd mostly failed to guide the swarm to the goal when using the method derived from~\cite{Strombom2014}, whereas the shepherd always succeeded in guiding the swarm when using the furthest agent targeting technique. The center-targeting method was found to perform similarly for both the global and local perspective simulations. 

%

\section{Alternative Shepherding Behaviors} \label{sec:alt}

While swarm shepherding originated from the herding of sheep, it has since been applied to numerous other types of tasks. Lien et al.~\cite{Lien2004} introduced four behaviors for which shepherding, as a swarm control technique, is applicable: \textit{herding} (classic shepherding), \textit{covering}, \textit{patrolling}, and \textit{collecting}. Covering is the process of guiding a swarm of agents to every location within an environment. Patrolling behavior involves a shepherd preventing swarm agents from entering a given space. Collecting sees the shepherd gather swarm agents, scattered around an environment, into a specific destination. Lien et al.~\cite{Lien2005} then conducted another experiment which investigated the benefit of including multiple shepherds in the swarm agent herding task. Results showed that a greater number of shepherds led to increased speed at guiding agents to an objective location, with the biggest difference in improvements seen between one and two shepherds. When the separation tendency of the swarm agents was intensified, however, the greatest increase in performance was observed between two and three shepherds. 

A new class of the shepherding problem, identified by Masehian and Royan~\cite{Masehian2015} as \textit{Simultaneous Object Collecting and Shepherding} (SOCS), which involves both the collecting and herding behaviors described in~\cite{Lien2004}. The authors developed a heuristic which reduces the complexity of the problem, while permitting online obstacle avoidance, incorporating swarm splitting, merging, deformation, and expansions behaviors, and uses a fuzzy decision model for environment exploration. Results showed that the decision model was not able to effectively trade-off time and number of agents collected and shepherded to the goal, as most simulations were not completed within the enforced time constraint. The heuristic was, however, able to shepherd the collected swarm online through complex obstacle-filled environments. 

While most research into shepherding has chosen to focus on the small-scale interactions between a shepherd and a swarm, one study by Cimler et al.~\cite{Cimler2016} developed an algorithm to investigate the large-scale shepherding behavior of animals belonging to ancient Celtic societies. As such, the simulated environment size was approximately $70km^2$, where different animal herds would graze on lands until the food source was depleted, then move to a different pasture. Further, each simulation was run for 40 years, where each time step represented a day and, thus, the motion of the animal agents was only taken as a daily displacement. Conversely to other studies, the shepherd in Cimler et al.\textquoteright s study would lead and the swarm would follow. Modification of the study to fit shorter timelines would be required to asses its application to the field of robotic swarm control, unless designed for a very large-scale project where only the daily displacement of the robotic system impacted their objective, thus real-time swarm dynamics are negligible.   

\section{Robotic and UxV Shepherding} \label{sec:rob}

\subsection{Robotic Shepherding} 

As robotics is one of the predominant goals of swarm shepherding as a control technique, several experimental studies have been conducted into its suitability for a wide variety of operations. One study by Chaimowicz and Kumar~\cite{Chaimowicz2007} used multiple uninhabited aerial vehicles (UAVs) to shepherd a swarm of uninhabited ground vehicles (UGVs), where the UGVs exhibited two main behaviors, \textit{splitting} and \textit{merging}. Splitting involved the swarm splitting into sub-groups, while merging was the converse. Each sub-group was assigned a shepherd, permitting a near-3D image of the environment to be obtained via a combination of airborne and ground-based UxV-mounted cameras. The objective of the swarm was to explore unknown terrains. Therefore, the splitting and merging behaviors were combined with the visual input, permitting the UGVs to effectively traverse complex environments. However, the methodologies and discussion of results are limited.

Razali et al.~\cite{Razali2013} conducted a study which focused on the identification of a robotic flock from other robotic agents within the same environment by using an image processing technique, called \textit{connected-components labeling}. Their results showed reasonable accuracies for independent flock recognition ($>70\%$); however, little information is provided regarding their simulated setup, and it is unclear if their algorithm would function on systems with greater complexity.

The ability to model the shepherding problem with simple systems is shown clearly in \"{O}zdemir, Gauci and Gro\textbeta \textquoteright s work~\cite{Ozdemir2017} on shepherding with simulated e-puck robots with no computation or memory capacity. The only information provided to the shepherds was from a single line-of-sight sensor which could differentiate between swarm agents, other shepherds, or the goal destination. This information was then translated into direct motor commands. The swarm agents, however, could only sense the shepherds and other agents. A reactive controller was optimized using an evolutionary robotics approach, such that the control parameters were modified to maximize a fitness function, designed to assess the performance of the shepherding task. The results showed that the shepherds often encircled the swarm agents, rotated in a clockwise direction around them, and then slowly moved in the direction of the goal.

A heuristic for the control of a swarm of uninhabited surface vessels (USVs) was developed by Long et al.~\cite{Long2019}. A UAV shepherd was to guide the USV swarm to a series of updating goal locations within a body of water in order to determine its wave properties in the local environment. The USV swarm move about the environment being repelled by each other, the shepherd and obstacles, and attracted by goal locations. Long et al.~\cite{Long20192} then ran a number of 2D simulations of a swarm being shepherded to a series of goal locations within an environment. The discrete forces applied in most previous studies (such as~\cite{Strombom2014}) were replaced with a continuous force, modulated with respect to radial distance from the source. A local environment path covering method, called a \textit{Hilbert space-filling curve} (HSFC), was introduced as a behavior, where swarm agents that strayed beyond a certain distance from the rest of the swarm began following the HSFC in search of goals. Further, a circular path planning algorithm was implemented by the shepherd to avoid unwanted disturbance of the swarm before the shepherd reached its target position. Results comparing traditional shepherding with the addition of the HSFC behavior showed that it was more efficient at reaching all the goals, and more effective at reaching a higher density of goals, within an environment. 

\subsection{Animal Shepherding} 

The application of robots to the actual act of shepherding a herd of sheep has also received attention~\cite{Evered2014},~\cite{Bat-Erdene2017}. Evered, Burling and Trotter~\cite{Evered2014} conducted a series of experiments using a three-wheeled-robot called SCRUPAL, which acted as a shepherd for 30 sheep on a 2.5 hectare paddock. The purpose of the experiments was to evaluate the reaction distance of the sheep when approached by SCRUPAL. The results showed that the sheep were initially repulsed by the robot at a distance of $60m$; however, after only two further trials, the repulsion distance had dropped to $10m$. This indicated that the sheep were becoming accustomed to the robot, and did not react to SCRUPAL in the same manner as they would to a sheepdog. No general conclusions could be made from the study, however, as the experimental results were limited, with only a single type of robot tested.

Pierson and Schwager~\cite{Pierson2015} discussed the idea of implementing their control strategy (discussed in Section~\ref{sec:control}) in UAVs for large-scale cattle herding, replacing the current piloted helicopters which are used for the task in order to reduce the significant casualty rates experienced~\cite{ATSB2017}. \cite{Pierson2015} implemented their shepherding design using Pololu m3pi robots which replicated multiple sheepdogs exhibiting herding behaviors, with similar experiments given in~\cite{Pierson2018}, where three shepherds were used to guide a single agent. Pierson and Schwager were successfully able to guide the swarm agent to a goal location, even though the inherent uncertainties and errors present in their experiments were discussed, highlighting the robustness of their control system. 

A control algorithm was developed by Bat-Erdene and Mandakh~\cite{Bat-Erdene2017} to investigate the potential for multiple robots to shepherd livestock. Two types of robots were used for their study: \textit{corner mobile robots} and \textit{sideline mobile robots}. Together, the robots were able to create square perimeters around the livestock of variable size. As such, the robots could enclose a swarm, transport it to another pasture to graze, enlarge their perimeter to the size of the pasture, then shrink the enclosure again when the livestock were required to move elsewhere. Bat-Erdene and Mandakh\textquoteright s paper created a framework for livestock shepherding; however, did not include any experimental or simulated results, thus, the algorithm was not validated.  

Gade, Paranjape and Chung~\cite{Gade2015} created an algorithm designed for a UAV to safeguard an airport\textquoteright s airspace from avian intruders by repelling them away, called the \textit{n-wavefront algorithm}. The n-wavefront algorithm is an example of the patrolling shepherding behavior described in~\cite{Lien2004}. Their research was motivated by the immense financial and temporal costs incurred by the aviation industry from wildlife collisions with aircraft~\cite{FAA2019}. The UAV shepherd aimed not only to guide the flock of birds, via its center of mass, but also to control the size of the flock. In order to achieve this, the shepherd selects \textit{n} birds on the boundary of the swarm to guide to ensure it maintains its shape (within the given bounds). A weighting was placed on the importance of maintaining the swarm shape versus guiding the swarm to a goal. Case studies where 100$\%$ boundary keeping, 100$\%$ goal seeking, and an intermediate 50/50 case, in both 2D and 3D environments, were compared. The results showed that the n-wavefront algorithm was better able to divert the bird flock away from the airport area than simply trying to guide them via their center. Further, the shape of the flock was better able to be maintained in 3D than in 2D. However, discussion of the results was quite limited. A follow-up study by Gade, Paranjape and Chung~\cite{Gade2016} aimed to analyze the communication model within a flock of birds, as well as focus on the stability of the n-wavefront algorithm. Results showed that modeled bird flocks with both first and second order dynamics, as well as directed star/tree communication topologies, converged to a formation exponentially quickly. Further, using the derived parameterization, the n-wavefront algorithm was found to be successful during 2D simulation. Continuing the research of Gade, Paranjape and Chung, Paranjape et al.~\cite{Paranjape2018} developed a different algorithm, called the \textit{m-waypoint algorithm}, for the same purpose. The algorithm was designed such that the UAV shepherd moves between \textit{m} different positions, randomly sampled in a distribution around the flock with a minimum distance separating them. The points were periodically regenerated in order to guide the swarm to the desired location, while maintaining the swarm shape. A series of experiments were conducted using a quadrotor drone shepherd, and two different flocks: egrets and loons. The egrets diverted their flight paths horizontally when approached at a sufficient distance by the UAV, whereas, when it came too close to the flock, they diverted vertically (continuing towards the airport\textquoteright s airspace). The UAV shepherd was also able to laterally deflect the faster loon flock, which was found to be comprised of three sub-flocks, from their flight path towards the airspace. The experiments, along with supporting simulations, showed the m-waypoint algorithm to be effective at shepherding bird swarms away from a protected airspace. 
 
\section{Human-Shepherding Interactions} \label{sec:hum}

A further area of interest in the field of robotics and swarm control is the interaction and collaboration of humans and machines. Lien and Pratt~\cite{Lien2009} developed a human-machine shepherd motion planning technique in order to augment their original roadmap-based control system~\cite{Bayazit2002} to overcome previously discussed challenges with shepherding control. The authors projected their shepherding simulation on a surface, then captured the resulting image using a camera. A user interacted with the simulation using laser pointers to alter the shepherding agents motion, resulting in real-time roadmap updates. The results showed that the human-machine interaction generally outperformed only-human or only-computer simulations, with respect to time. 

Nalepka et al.~\cite{Nalepka2017} ran a series of experiments involving a two-player shepherding game, designed to investigate human behavioral dynamics during a complex, multi-agent task. The objective of the game was to guide a number of swarm agents into a central circle, within a bounded environment, and keep them all there 75$\%$ of the trial time (60 seconds). The task was repeated a certain number of times, or until 45 minutes was exceeded. The swarm moved about the environment via Brownian motion, and if a swarm agent collided with the boundary, then that trial was failed. Further, the players were not able to verbally communicate with each other before or during the trials, merely having to discover a method of cooperating in-game. The vast majority of pairs utilized two strategies to keep the swarm agents centered: \textit{search-and-rescue} (S$\&$R) or \textit{coupled-oscillatory-containment} (COC). S$\&$R saw players each collect the furthest agent in one half of the field, which, after a variable amount of trials, often led to COC where the players would oscillate in a semi-circular pattern on either side of the central circle to contain the agents inside. Nalepka et al.~\cite{Nalepka2016} then created an \textit{expert artificial agent} (EAA) designed to behave as the human subjects did in~\cite{Nalepka2017} (S$\&$R and COC), with the objective of modeling the human behaviors, rather than trying to optimize the shepherding task. Human players were then randomly assigned to play the shepherding game with either other human players or the EAA. In this study, the game was played in virtual reality, and at completion, participants were asked if they suspected that they were playing with another human. Twelve out of 18 players were convinced by the EAA, believing that they were playing with a human. In this study, however, the EAA automatically switches to COC once a certain level of proficiency is reached, which led to human players mimicking the EAA rather than discovering COC on their own. Therefore, Nalepka et al.~\cite{Nalepka2018} conducted a separate study, in which the EAA was not able to automatically switch from S$\&$R to COC. As the EAA was incapable of the oscillatory motion achieved by humans in~\cite{Nalepka2017} and humans and machines in~\cite{Nalepka2016}, the exact behavior was not replicated. However, some of the players created a slower alternative, where they would complete semi-circular oscillations and the EAA would collect the agents on the other half of the environment, indicating that the participants were capable of discovering the behavior without the need for a partner. 
Taking the concept a step further, Nalepka et al.~\cite{Nalepka2019} designed a virtual reality game where the players acted as the shepherds on the field by physically moving about a $6m\times3.48m$ environment. The participants in this experiment exhibited behavior similar to that of the 2D hand-controlled game. At first the players would collect the furthest agents from the centre (S$\&$R). As the speed of the agents increased, the players would then often start circling the agents to contain them in the centre (rather than oscillate across a semicircle). This was more efficient as it did not require the participants to change direction, while during the hand-controlled task, circular rotation would have required participants to cross arms.

A multi-robot system, devised by Tuyls et al.~\cite{Tuyls2016}, introduced a telepresence robot, operated by a human, to augment a swarm robotic foraging task based on insect foraging behaviors. The human operator acted as a shepherd, controlling a MITRO robot to help steer the robotic swarm towards \textit{food sources} at specific locations within an environment, or to collect lost agents. The objective of the robots was to collect all of the food and return it to a \textit{hive} location, which was done by randomly searching for food initially, then transmitting the known source locations to other passing robotic agents until they were all foraging between the hive and food sources. Results from both simulations and actual experiments showed that the telepresence robot shepherd augmented the foraging task when compared with a shepherdless, honeybee-based foraging swarm, particularly as the complexity of the task increased.

Wade and Abbass~\cite{Wade2019CyberShepherd} designed a shepherding game they called \textit{Cyber-Shepherd}. The basic motivation of the game is to create a data collection environment for machine learning agents to learn from humans playing the game. The game had 15 levels of difficulty, and was designed to be suitable to operate on a smart phone and to provide data for supervised and reinforcement learning algorithms similar to those presented in~\cite{alex2019machine},~\cite{clayton2019machine}, in Section~\ref{sec:ML}. While Wade and Abbass~\cite{Wade2019CyberShepherd} relied on simple image-base visualization of the sheep and dog, Skrba et al.~\cite{Skrba2006} present a system for animation and rendering of sheep and sheepdogs. Their emphasis was mainly placed on animation with no modeling or discussion of the sheep-sheepdog interaction.

\section{Machine Learning for Smart Shepherding} \label{sec:ML}

The use of machine learning and computational intelligence techniques for swarm control~\cite{Nguyen2018QLearning} in general, and for shepherding in particular, is still in its infancy. The learning problem is non-trivial due to the large search space~\cite{Lien2009} for a machine learner that can generalize well. This has motivated an area of research into the adoption of human education methodologies into machine education.

Machine teaching attempts to structure a task for a machine learner in incrementally designed sub-tasks. Elman~\cite{Elman1993} is normally attributed with the introduction of the concept, where he sequenced the learning task by incrementally increasing the complexity of the concepts to be learnt, while simultaneously increasing the computational resources of the learner. However, machine teaching does not delve into pedagogical questions such as: how to structure the learning experience for a machine learner to learn complex skills. This question is central in machine education. Abbass et al.~\cite{Abbass2019} laid out the pedagogy of this process.

Gee and Abbass~\cite{alex2019machine} used machine education to teach supervised neural networks how to shepherd. They systematically divided the shepherding task into its driving and collecting sub-skills, then used a human to form a database of demonstrations for each sub-skill. An advantage of their approach is the low cost associated with collecting data from humans compared to setting up appropriate experiments with real sheep. The latter would be very difficult to achieve. For example, to generate a dataset where a group of sheep is clustered 90 degrees from the goal and a shepherd positioned 270 degrees from the goal, using real sheep and sheepdogs, is very complex. Such a task can be easily setup in a simulation environment, or using physical robots in a physical environment, in a much simpler manner. Gee and Abbass~\cite{alex2019machine} demonstrated that the neural network successfully learnt the sub-skills using the human demonstrations.

Clayton and Abbass~\cite{clayton2019machine} used machine education with a slight variation from the pedagogical framework used by Gee and Abbass~\cite{alex2019machine}, for reinforcement learning (RL) agents. While the representation was still a neural network, they did not rely on human demonstrations, instead attempting to use RL to learn each sub-skill independently. A primary challenge for an RL agent is the design of the reward function. Clayton and Abbass used a machine education framework to design the reward function using a decomposition approach similar to the design of informative and summative assessments in human education. The research illustrated that machine education structures the reward function better than attempting to learn shepherding as a single task.

Nguyen et al.~\cite{Nguyen2018Apprenticeship},~\cite{Nguyen2018Apprenticeship2}, designed an approach they called \textit{Apprenticeship Bootstrapping} (AB), where the task is decomposed into sub-skills. The decomposition needs to continue until there are human experts who can perform the individual sub-skills properly. The AB approach uses the machine learning models learnt from human demonstrations on these sub-skills to aggregate a model on the larger task. Nguyen et al.~\cite{Nguyen2018Apprenticeship},~\cite{Nguyen2018Apprenticeship2}, initially tested their methodology using a leader-follower setup, where a UAV aims to follow a swarm of ground vehicles. The UAV used AB to learn how to follow and maintain the whole swarm within its field of view at all times.

The AB approach was then extended by Nguyen et al.~\cite{Nguyen2019ICONIP} to shepherd the swarm. The task was decomposed into the two basic skills of collecting and driving, followed by the use of a human to perform each task independently. Inverse reinforcement learning and deep reinforcement learning were then used to transform the set of demonstrations on each sub-skill into a model, then AB was used to aggregate these models into an overall shepherding UAV.

Two further examples of machine learning for shepherding are worth noting. The first focuses on the use of genetic algorithms (GAs) to evolve the parameters of a shepherding model. Singh et al.~\cite{Singh2019} used a GA to optimise the level of modulation of the force vector exerted by the sheepdog on the sheep. In practice, this force vector is correlated with the fuel/energy needed by the sheepdog. The modulation of the force vector aims at ensuring that the force to be exerted on the sheep is at the right level to achieve the intent. If this force is too low, the sheep will not respond as intended. If this force is too high, the sheep will scatter and the artificial or biological sheepdog will waste energy. Singh et al. demonstrated that optimising the modulation function using a GA saves energy, although the fixed model for shepherding used in the experiments constrained the optimization.

The second example of research on machine learning and shepherding was conducted by Abbass et al.~\cite{Abbass2018}, where the authors used the concept of shepherding for AI testing. The complexity of the AI within an autonomous system will continue to increase with no adequate methodologies to test these algorithms in advance due to the enormous search space they operate within, the dynamic nature of the environment, and their learning abilities that change their algorithmic performance based on the experience they encounter. Abbass et al.~\cite{Abbass2018} designed a concept whereby an \textit{AI Shepherd} is used to shepherd a group of AI (artificial sheep) by continuously monitoring the group and applying corrective actions as needed. This concept offered what the authors called \textit{Watchdog AI} (WAI) agents, where each WAI is basically an AI shepherd. The authors presented their framework theoretically and listed the specifications of a WAI. However, they did not demonstrate the concept with a case study due to the complexity of implementing a concept of this nature. 

\section{Shepherding Challenges} \label{sec:chal}

\begin{figure*}[!t]
\centering
\includegraphics[scale=0.73]{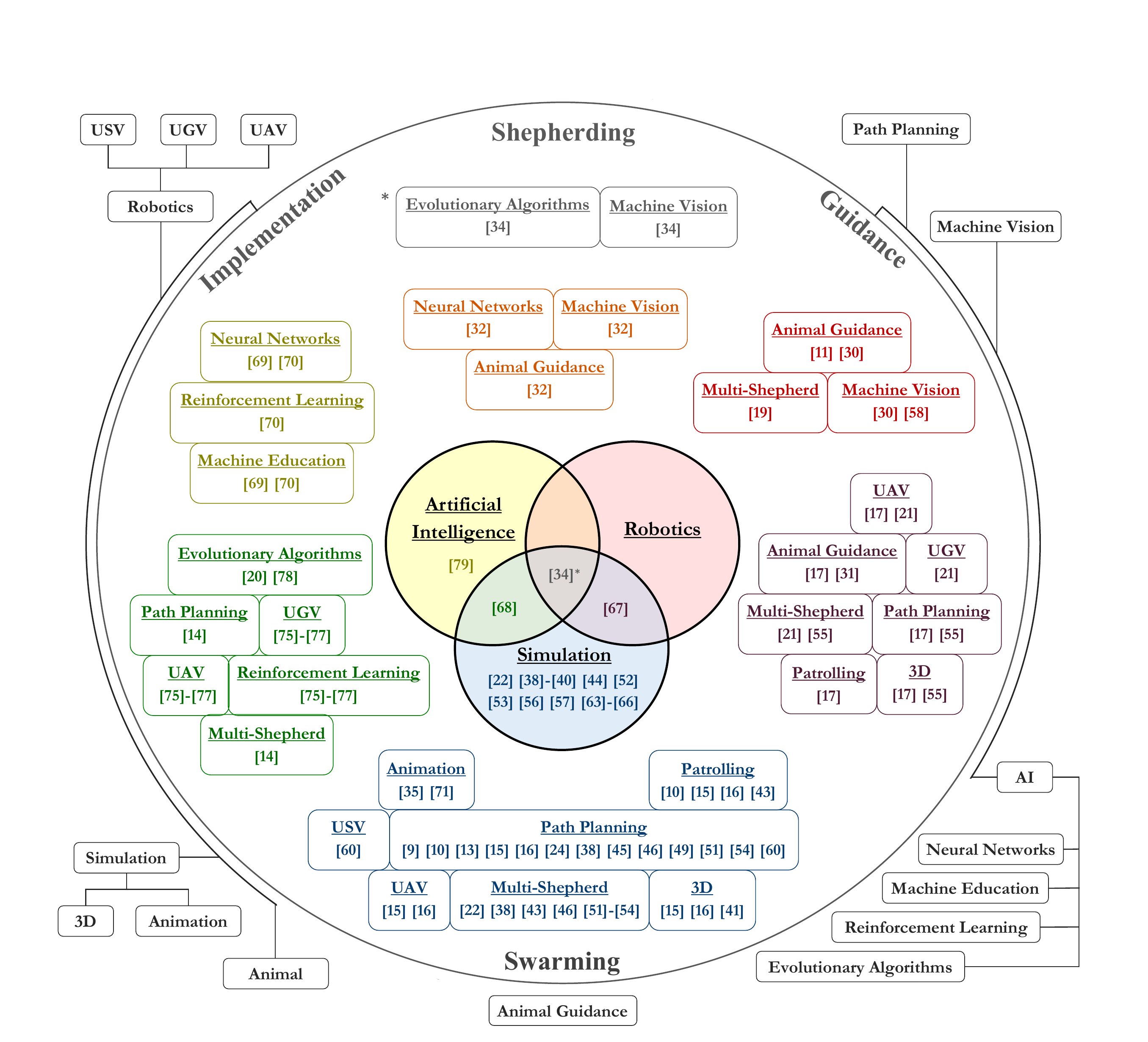}
\caption{Shepherding mind map. The inner circle is used to categorize the literature as simulation, robotics and artificial intelligence, or a combination of them. Each of these categories is then decomposed into associate sub-categories. A hierarchical tree of shepherding taxonomy is then presented outside the circle, initially split into shepherding guidance or implementation.}
\label{fig:shep_diag}
\end{figure*}

		\begin{table}
		    \centering
				\caption{Shepherding glossary.}
				\label{tab:shep_glossary}
				\begin{tabularx}{\linewidth}{| s | g |}
				    \hline
				    \textbf{Term}	& \textbf{References} \\ \hline
		            {3D} &	[15]-[17] [41] [55] \\ \hline
                    AI &	[14] [20] [32] [34] [68]-[70] [75]-[79] \\ \hline
                    Animal Guidance &	[11] [17] [30]-[32] \\ \hline
                    Animation &	[35] [71] \\ \hline
                    Evolutionary Algorithms &	[20] [34] [78] \\ \hline
                    Machine Education &	[69] [70] \\ \hline
                    Machine Vision &	[30] [32] [34] [58] \\ \hline
                    Multi-shepherd &	[14] [19] [21] [22] [38] [43] [46] [51]-[55] \\ \hline
                    Neural Networks &	[32] [69] [70]  \\ \hline
                    Path Planning &	[9] [10] [13]-[17] [24] [38] [45] [46] [49] [51] [54] [55] [60] \\ \hline
                    Patrolling &	[10] [15]-[17] [43] \\ \hline
                    Reinforcement Learning &	[70] [75]-[77] \\ \hline
                    Robotics &	[11] [17] [19] [21] [30]-[32] [34] [55] [58] [67] \\ \hline
                    Simulation &	[9] [10] [13]-[17] [20]-[22] [24] [31] [34] [35] [38]-[41] [43]-[46] [49] [51]-[57] [60] [63]-[68] [71] [75]-[78] \\ \hline
                    UAV &	[15]-[17] [21] [75]-[77]  \\ \hline
                    UGV &	[21]  [75]-[77] \\ \hline
                    USV &	[60] \\ \hline
                \end{tabularx}
			
		\end{table}

A mind map of shepherding that has been developed to dissect the interconnected relationships between key shepherding elements, is presented in Figure \ref{fig:shep_diag}. Firstly, the literature within the circle is related to AI, robotics, simulation, or any of the combinations of the three (color coordinated). Within this decomposition, the associated recurring elements are presented with the relevant literature. This decomposition highlights unique common features between the shepherding research presented in this review. The highly populated bottom half of the figure reveals that the majority of research has involved some form of simulation, whether or not it also incorporates robotic experiments or AI techniques. The studies which involved some form of path planning, beyond simply reacting to influential force vectors, then form the largest categorization within the structures presented. Patrolling was found to be the second most common shepherding objective after herding (not included in figure). Another interesting feature was the widespread use of UxVs for shepherding tasks, where UAVs were the most common. The secondary analysis presented in Figure \ref{fig:shep_diag} was the hierarchical categorization of shepherding taxonomy based on its association to either shepherding guidance or the implementation of shepherding (outside of the circle). Animal guidance is the only term floating outside of the tree as it is associated with both guidance and implementation. Shepherding and swarming form the core base of shepherding research, therefore, they act as the roots from which all of the survey elements in Figure \ref{fig:shep_diag} emanate.

Table \ref{tab:shep_glossary} provides a summary of the key shepherding elements presented in Figure \ref{fig:shep_diag}, acting as a glossary of the research presented in the review. Therefore, Table \ref{tab:shep_glossary} can be used as a direct reference for the main trends in shepherding, while Figure \ref{fig:shep_diag} shows an expanded web of shepherding relationships that can be used to extract specialised references.

While shepherding as a swarm control technique has matured, there remain a plethora of challenges. These have been broken down into challenges arising from modelling, the role of machine learning and artificial intelligence, issues directly related to shepherding in robotics, as well as the analytics of shepherding.

\subsection{Modelling}

The majority of the literature reviewed in this study relied on a human to infer a model and a set of rules that could be used by an artificial agent to produce shepherding behaviors. This presents a number of limitations. 

Firstly, human perception can be biased. For example, Harvey, Merrick and Abbass~\cite{Harvey2018} identified regions in the parameter space for a boid model where swarming occurs according to mathematical metrics, while humans were unable to detect these regions. They demonstrated characteristic differences in the perception of swarming. If humans are unable to see the full spectrum of swarming, one could wonder if models driven by human perception of swarming are biased or incomplete.

Secondly, there is no guarantee that the human-designed models discussed here would be the most efficient way to drive an artificial agent to shepherd, nor that it is the right and/or only way. A robotic swarm has constraints on memory, processing resources, and energy. The transfer of behavior from, for example, sheepdogs to a robot does not entail that the behaviour is effective and/or efficient for the robot in the same manner that it was suitable for a sheepdog. Similarly, a collection of UAVs shepherding collectively, for example, have different physical constraints from a group of biological sheepdogs.

Thirdly, the models reviewed here are not easily adapted in a dynamic environment and they usually cannot generalize to other forms of shepherding. The literature review up to this section has demonstrated the effort that goes into transferring a model to a different context, or even the effect of a parameter, such as swarm size on one model~\cite{Hoshi2018},~\cite{Hoshi20182}.

Additionally, the above work on machine teaching, especially by Gee and Abbass~\cite{alex2019machine}, relies on the availability of a human expert to collect demonstrations. Human experts are expensive and their time is normally limited, representing a significant limitation for the collection of sufficiently diverse and large number of demonstrations for a machine learner to learn from. Moreover, assuming a complex task such as swarm guidance, the problem itself is that it is an emerging technology with no existing standards on how to perform the task, let alone the existence of a human to claim expertise in performing the task.

\subsection{Intelligence}
Another perspective on this problem is to ask whether the artificial agent itself can discover these rules or other rules for shepherding on its own. There are many advantages to explore this line of research. However, this literature review revealed that this line of research is still at its infancy. Many opportunities exist for computational intelligence and machine learning algorithms in shepherding. Three proposed examples are outlined below.

Firstly, machine learning techniques to dynamically learn the behaviour of agents in a swarm, in order to determine and modulate an appropriate set of characteristics for the influence vector, would be extremely useful due to the variations of behaviors that a shepherd could be faced with in the real world. For example, in the case of using a robotic shepherd in the actual sheep herding task, different sheep exhibit different behaviors depending on their behavioral profile within the same breed or across different breeds. For the shepherd to be useful from an industry perspective, it needs to be able to learn these behavioral profiles and calibrate its decision making model accordingly. This also calls for the need to use real-time optimization where the shepherd is able to optimize its behaviour as the context changes. 

Secondly, all models explored in the above literature rely on taking a weighted sum of the forces. Finding the right set of weights is a context dependent problem. These weights may need to vary as the number of agents in the swarm, the number of shepherds, or the terrain is modified, or if the type of information and level of uncertainty from the sensors change. This level of complexity requires an appropriate AI for planning and adapting the shepherd\textquoteright s behavior in dynamic and uncertain environments.

Thirdly, the large role a handler plays when guiding a sheepdog lies in the handler\textquoteright s ability to recognize the activities of a herd of sheep at a particular point of time. For example, if a sheep just started to eat, herding may need to be rescheduled to allow the sheep to obtain sufficient energy for the journey ahead. If a group of sheep splits from the swarm, running towards an area with possible threats such as an area of foxes, then the handler may need to command the sheepdog to start by collecting these sheep. When this intelligence is transformed into an artificial sheepdog, activity recognition becomes a vital problem that needs to be resolved. While there is significant literature on computer-based activity recognition in the human and information world, little research has taken place on this topic in the animal world.

\subsection{Robotics}

When focusing on the multi-agent and robotics context, three main challenges are worth discussing.

Firstly, in the case of a single shepherd, the control system cannot be considered entirely decentralized as one entity processes all of the high-level planning. As such, though the swarm itself remains redundant, the malfunctioning of the shepherd could result in a total system failure. This can be mitigated with the introduction of multiple shepherds, as long as each is capable of operating independently of each other. 

Secondly, while in principle the failure of a single agent should not lead to the failure of the swarm as a whole, this has not been taken into account in all of the literature so far. For example, if a shepherding algorithm was implemented using robotic agents, if one robotic agent was to break down, the shepherd would continually try to collect it back into the swarm without success, thereby never reaching the objective destination. An example of a simple solution to this problem would be to include a condition in the program which ignores agents which have not displaced over a given number of time steps, although, more complex solutions would likely be required.

Thirdly, \c{C}elikkanat and \c{S}ahin~\cite{Celikkanat2010} discuss the impact of self-organizing swarms, which rely on local interactions, leading to difficulties in fully controlling the system. While the shepherd may be capable of influencing a swarm to its advantage, the nature of the swarm itself makes complete control of it onerous, limiting the practical applications to those that do not require precise dynamics. Further, in the case of robotic systems, this would lead to non-optimal time duration and energy consumption. Therefore, the trade-off between robustness and efficiency must be considered before employing swarm shepherding, or even swarming in general, for a task. It should be recognized, however, that controlling a group of animals which do not exhibit swarming behaviors (such as cats) would be far more difficult than those which do~\cite{Bennett2012}.

\subsection{Analytics}
Finally, there is little research on tools to analyze the interaction space between the sheep, between the shepherds and between the sheep and shepherds. Research into social network analysis could be very useful here. Similarly, the influence vectors among the sheep and the shepherds are in effect messages that the agents are either sensing or exchanging. Understanding this implicit communication network could reveal new forms of communication and better understanding of how to optimize the shepherding system. Moreover, the work of Tang et al.~\cite{tang2017networking} demonstrated that explicit networking of the agents make it more difficult for an adversary to reverse engineer and attack the dynamics of the agents. This raises the question of whether or not there are similar advantages in networking the shepherds, robotic or simulated. One main advantage of shepherding is the absence of an explicit network. Therefore, the pros of explicit networking need to be weighed against the cons. 

\subsection{Future Shepherding}

The challenges outlined in the previous subsections present the core of the research on shepherding to follow. This could be complemented by an expansion upon current planning methods including path and goal planning. Moreover, the behavioral set of the swarm, sheepdogs and shepherds that have been covered so far in the literature, seem to be limited to very primitive behaviours. Increasing the richness of the behavior set will increase the complexity of the expressed dynamics. Adding more complex AI to support the shepherding task will equally make these systems exhibit richer forms of natural intelligence.

The various branches of shepherding presented in this review will also likely intertwine and combine in the future. For example, taking the case of sheep herding using robotic agents, the future could see a hierarchical system implemented which involves multiple UxV shepherds guiding the sheep to an objective location, and a human in the loop monitoring the task, with the ability to actively modify the shepherds\textquoteright behaviors and trajectory remotely and in real-time.

It should also be noted that additional advancements in one of the shepherding sub-fields could be directly applied to advancements in another. For example, the development of an advanced sheep herding robotic shepherd could lead to its modification and use as crowd control robot, where the reverse could be equally as true.

\section{Conclusion}

Over 30 years, a large body of research has been accumulated on the use of shepherding as a swarm control technique. The aggregate of results show great promise for the application of shepherding to swarm systems, enabling robust, flexible, scalable yet simple solutions to complex tasks, where the low-level control is designated to the swarm, and high-level planning is assigned to the shepherd(s).

While the majority of experimentation has been simulation, the eventual outcome is generally intended for robotic systems. Examples of robotic applications have included navigating uninhabited ground vehicles through complex terrains, distributing uninhabited surface vehicles for the purpose of sea state estimation, using robotic shepherds to herd sheep, and guiding bird swarms away from airports using uninhabited aerial vehicles, amongst several others.

Despite the considerable range of literature to date on swarm shepherding, there still remains a broad scope of research areas for more studies into a wider range of applications. We have laid out key research gaps and challenges in the domain of shepherding, that range from machine learning and artificial intelligence, to robotic and multi-agent systems. 
None of these challenges are insurmountable, leaving the future of swarm shepherding full of potential and opportunities to be seized by researchers and technologists.

\section*{Acknowledgment}
This research is conducted in collaboration with the Maritime Division of the Defence Science and Technology Group and is partially supported by the Office of Naval Research - Global Grant N62909-18-1-2140-P00001.

\ifCLASSOPTIONcaptionsoff
  \newpage
\fi

\bibliographystyle{IEEEtran}

\bibliography{ShepherdingReviewPaper}

\begin{thebibliography}{10}
\providecommand{\url}[1]{#1}
\csname url@rmstyle\endcsname
\providecommand{\newblock}{\relax}
\providecommand{\bibinfo}[2]{#2}
\providecommand\BIBentrySTDinterwordspacing{\spaceskip=0pt\relax}
\providecommand\BIBentryALTinterwordstretchfactor{4}
\providecommand\BIBentryALTinterwordspacing{\spaceskip=\fontdimen2\font plus
\BIBentryALTinterwordstretchfactor\fontdimen3\font minus
  \fontdimen4\font\relax}
\providecommand\BIBforeignlanguage[2]{{%
\expandafter\ifx\csname l@#1\endcsname\relax
\typeout{** WARNING: IEEEtran.bst: No hyphenation pattern has been}%
\typeout{** loaded for the language `#1'. Using the pattern for}%
\typeout{** the default language instead.}%
\else
\language=\csname l@#1\endcsname
\fi
#2}}

\bibitem{Zoghby2013}
N.~E. Zoghby, V.~Loscrí, E.~Natalizio, and V.~Cherfaoui, \emph{Chapter 8:
  Robot Cooperation and Swarm Intelligence}.\hskip 1em plus 0.5em minus
  0.4em\relax World Scientific, 2013, pp. 163--201.

\bibitem{Harvey2018}
J.~Harvey, \emph{The Blessing and Curse of Emergence in Swarm Intelligence
  Systems}.\hskip 1em plus 0.5em minus 0.4em\relax Cham: Springer International
  Publishing, 2018, pp. 117--124.

\bibitem{Parker1998}
L.~E. {Parker}, ``{ALLIANCE}: an architecture for fault tolerant multirobot
  cooperation,'' \emph{IEEE Transactions on Robotics and Automation}, vol.~14,
  no.~2, pp. 220--240, Apr 1998.

\bibitem{Gazi2011}
V.~Gazi and K.~M. Passino, \emph{Swarm Coordination and Control
  Problems}.\hskip 1em plus 0.5em minus 0.4em\relax Berlin, Heidelberg:
  Springer Berlin Heidelberg, 2011, pp. 15--25.

\bibitem{Gazi2005}
V.~{Gazi}, ``Swarm aggregations using artificial potentials and sliding-mode
  control,'' \emph{IEEE Transactions on Robotics}, vol.~21, no.~6, pp.
  1208--1214, Dec 2005.

\bibitem{Gazi2012}
V.~Gazi, B.~Fidan, R.~Ord\'{o}\~{n}ez, and M.~\.{I}lter K\"{o}ksal, ``{A Target
  Tracking Approach for Nonholonomic Agents Based on Artificial Potentials and
  Sliding Mode Control},'' \emph{Journal of Dynamic Systems, Measurement, and
  Control}, vol. 134, no.~6, Sep 2012.

\bibitem{Lim2009}
H.-C. Lim and H.~Bang, ``Adaptive control for satellite formation flying under
  thrust misalignment,'' \emph{Acta Astronautica}, vol.~65, no.~1, pp. 112 --
  122, 2009.

\bibitem{Gazi2015}
V.~Gazi, B.~Fidan, L.~Marques, and R.~Ord\'{o}\~{n}ez, \emph{Robot Swarms:
  Dynamics and Control}.\hskip 1em plus 0.5em minus 0.4em\relax ASME, Jan 2015,
  pp. 79--125.

\bibitem{Vo2009}
C.~Vo, J.~F. Harrison, and J.-M. Lien, ``Behavior-based motion planning for
  group control,'' in \emph{IEEE/RSJ International Conference on Intelligent
  Robots and Systems}.\hskip 1em plus 0.5em minus 0.4em\relax IEEE, 2009, pp.
  3768--3773.

\bibitem{Lien2004}
J.-M. Lien, O.~Bayazit, R.~Sowell, S.~Rodriguez, and N.~Amato, ``Shepherding
  behaviors,'' in \emph{Proceedings of the 2005 IEEE International Conference
  on Robotics and Automation}, vol.~4, Jan 2004, pp. 4159 -- 4164 Vol.4.

\bibitem{Evered2014}
M.~Evered, P.~Burling, and M.~Trotter, ``An investigation of predator response
  in robotic herding of sheep,'' \emph{International Proceedings of Chemical,
  Biological and Environmental Engineering}, vol.~63, pp. 49--54, 2014.

\bibitem{Bat-Erdene2017}
B.~{Bat-Erdene} and O.~{Mandakh}, ``Shepherding algorithm of multi-mobile robot
  system,'' in \emph{First IEEE International Conference on Robotic Computing
  (IRC)}, Apr 2017, pp. 358--361.

\bibitem{Lien2009}
J.-M. Lien and E.~Pratt, ``Interactive planning for shepherd motion.'' in
  \emph{AAAI Spring Symposium: Agents that Learn from Human Teachers}, Mar
  2009, pp. 95--102.

\bibitem{Masehian2015}
E.~Masehian and M.~Royan, \emph{Cooperative Control of a Multi Robot Flocking
  System for Simultaneous Object Collection and Shepherding}.\hskip 1em plus
  0.5em minus 0.4em\relax Cham: Springer International Publishing, 2015, pp.
  97--114.

\bibitem{Gade2015}
S.~Gade, A.~A. Paranjape, and S.-J. Chung, ``Herding a flock of birds
  approaching an airport using an unmanned aerial vehicle,'' in \emph{AIAA
  Guidance, Navigation, and Control Conference}, 2015, p. 1540.

\bibitem{Gade2016}
------, ``Robotic herding using wavefront algorithm: Performance and
  stability,'' in \emph{AIAA Guidance, Navigation, and Control Conference},
  2016, p. 1378.

\bibitem{Paranjape2018}
A.~A. {Paranjape}, S.~{Chung}, K.~{Kim}, and D.~H. {Shim}, ``Robotic herding of
  a flock of birds using an unmanned aerial vehicle,'' \emph{IEEE Transactions
  on Robotics}, vol.~34, no.~4, pp. 901--915, Aug 2018.

\bibitem{Shedied2013}
S.~A. Shedied, ``Optimal trajectory planning for the herding problem: a
  continuous time model,'' \emph{International Journal of Machine Learning and
  Cybernetics}, vol.~4, no.~1, pp. 25--30, Feb 2013.

\bibitem{Pierson2015}
A.~{Pierson} and M.~{Schwager}, ``Bio-inspired non-cooperative multi-robot
  herding,'' in \emph{IEEE International Conference on Robotics and Automation
  (ICRA)}, May 2015, pp. 1843--1849.

\bibitem{Ozdemir2017}
A.~\"{O}zdemir, M.~Gauci, and R.~Gro\textbeta, ``Shepherding with robots that
  do not compute,'' in \emph{The Fourteenth European Conference on Artificial
  Life}, no.~29, 2017, pp. 332--339.

\bibitem{Chaimowicz2007}
L.~Chaimowicz and V.~Kumar, ``Aerial shepherds: Coordination among {UAVs} and
  swarms of robots,'' in \emph{Distributed Autonomous Robotic Systems 6},
  R.~Alami, R.~Chatila, and H.~Asama, Eds.\hskip 1em plus 0.5em minus
  0.4em\relax Tokyo: Springer Japan, 2007, pp. 243--252.

\bibitem{Razali2012}
S.~Razali, Q.~Meng, and S.-H. Yang, ``Immune-inspired cooperative mechanism
  with refined low-level behaviours for multi-robot shepherding,''
  \emph{International Journal of Computational Intelligence and Applications},
  vol.~11, no.~01, p. 1250007, 2012.

\bibitem{IntSheepdogSoc2018}
\BIBentryALTinterwordspacing
{International Sheep Dog Society}, ``Rules for trials,'' 2018. [Online].
  Available:
  \url{https://www.isds.org.uk/trials/sheepdog-trials/rules-for-trials/}
\BIBentrySTDinterwordspacing

\bibitem{Bennett2012}
B.~Bennett and M.~Trafankowski, ``A comparative investigation of herding
  algorithms,'' in \emph{AISB/IACAP World Congress 2012: Understanding and
  Modelling Collective Phenomena}, Jan 2012.

\bibitem{Keil2015}
P.~G. Keil, ``Human-sheepdog distributed cognitive systems: An analysis of
  interspecies cognitive scaffolding in a sheepdog trial,'' \emph{Journal of
  Cognition and Culture}, vol.~15, no.~5, pp. 508 -- 529, 2015.

\bibitem{S1012019}
\BIBentryALTinterwordspacing
{Sheep 101}, ``Basic information about sheep,'' 2015. [Online]. Available:
  \url{http://www.sheep101.info/sheepbasics.html}
\BIBentrySTDinterwordspacing

\bibitem{Werner1993}
G.~M. Werner and M.~G. Dyer, ``Evolution of herding behavior in artificial
  animals,'' in \emph{From Animals to Animats 2: Proceedings of the Second
  International Conference on Simulation of Adaptive Behavior}, vol.~2.\hskip
  1em plus 0.5em minus 0.4em\relax MIT Press, 1993, p. 393.

\bibitem{coren2006intelligence}
S.~Coren, \emph{The intelligence of dogs: A guide to the thoughts, emotions,
  and inner lives of our canine companions}.\hskip 1em plus 0.5em minus
  0.4em\relax Simon and Schuster, 2006.

\bibitem{DHT2019}
\BIBentryALTinterwordspacing
{Doghouse Training}, ``{Working Dog Auction Results Held on Sunday, June 10th
  2018},'' 2018. [Online]. Available: \url{https://farmdogtraining.com.au/}
\BIBentrySTDinterwordspacing

\bibitem{Vaughan1997}
R.~Vaughan, J.~Henderson, and N.~Sumpter, ``Introducing the robot sheepdog
  project,'' in \emph{Proceedings of the International Workshop on Robotics and
  Automated Machinery for BioProductions}, Valencia, Spain, 1997.

\bibitem{Vaughan1998}
R.~Vaughan, N.~Sumpter, A.~Frost, and S.~Cameron, ``Robot sheepdog project
  achieves automatic flock control,'' in \emph{Proceedings of the Fifth
  International Conference on the Simulation of Adaptive Behaviour}, vol.
  489.\hskip 1em plus 0.5em minus 0.4em\relax 493, 489, 1998, p. 493.

\bibitem{Sumpter1998}
N.~Sumpter, A.~J. Bulpitt, R.~T. Vaughan, R.~D. Tillett, and R.~D. Boyle,
  ``Learning models of animal behaviour for a robotic sheepdog.'' in \emph{The
  International Association for Pattern Recognition Workshop on Machine Vision
  Applications}, Nov 1998, pp. 577--580.

\bibitem{Vaughan2000}
R.~Vaughan, N.~Sumpter, J.~Henderson, A.~Frost, and S.~Cameron, ``Experiments
  in automatic flock control,'' \emph{Robotics and Autonomous Systems},
  vol.~31, no.~1, pp. 109 -- 117, 2000.

\bibitem{Schultz1999}
A.~Schultz, J.~J. Grefenstette, and W.~Adams, ``Robo-shepherd: Learning complex
  robotic behaviors,'' in \emph{In Robotics and Manufacturing: Recent Trends in
  Research and Applications}, vol.~6.\hskip 1em plus 0.5em minus 0.4em\relax
  ASME Press, 1999, pp. 763--768.

\bibitem{Funge1999}
J.~Funge, X.~Tu, and D.~Terzopoulos, ``Cognitive modeling: knowledge, reasoning
  and planning for intelligent characters,'' \emph{Computer Graphics}, pp.
  29--38, 1999.

\bibitem{Kachroo2001}
P.~{Kachroo}, S.~A. {Shedied}, J.~S. {Bay}, and H.~{Vanlandingham}, ``Dynamic
  programming solution for a class of pursuit evasion problems: the herding
  problem,'' \emph{IEEE Transactions on Systems, Man, and Cybernetics, Part C
  (Applications and Reviews)}, vol.~31, no.~1, pp. 35--41, Feb 2001.

\bibitem{Kachroo2002}
P.~{Kachroo}, S.~A. {Shedied}, and H.~{Vanlandingham}, ``Pursuit evasion: the
  herding noncooperative dynamic game - the stochastic model,'' \emph{IEEE
  Transactions on Systems, Man, and Cybernetics, Part C (Applications and
  Reviews)}, vol.~32, no.~1, pp. 37--42, Feb 2002.

\bibitem{Miki2006}
T.~{Miki} and T.~{Nakamura}, ``An effective simple shepherding algorithm
  suitable for implementation to a multi-mobile robot system,'' in \emph{First
  International Conference on Innovative Computing, Information and Control -
  Volume I (ICICIC'06)}, vol.~3, Aug 2006, pp. 161--165.

\bibitem{Strombom2014}
D.~Str\"{o}mbom, R.~P. Mann, A.~M. Wilson, S.~Hailes, A.~J. Morton, D.~J.~T.
  Sumpter, and A.~J. King, ``Solving the shepherding problem: heuristics for
  herding autonomous, interacting agents,'' \emph{Journal of The Royal Society
  Interface}, vol.~11, no. 100, p. 20140719, 2014.

\bibitem{Hoshi2018}
H.~Hoshi, I.~Iimura, S.~Nakayama, Y.~Moriyama, and K.~Ishibashi, ``Robustness
  of herding algorithm with a single shepherd regarding agents\textquoteright \
  moving speeds,'' \emph{Journal of Signal Processing}, vol.~22, no.~6, pp.
  327--335, Nov 2018.

\bibitem{Hoshi20182}
H.~{Hoshi}, I.~{Iimura}, S.~{Nakayama}, Y.~{Moriyama}, and K.~{Ishibashi},
  ``Computer simulation based robustness comparison regarding agents'
  moving-speeds in two- and three-dimensional herding algorithms,'' in
  \emph{Joint 10th International Conference on Soft Computing and Intelligent
  Systems (SCIS) and 19th International Symposium on Advanced Intelligent
  Systems (ISIS)}, Dec 2018, pp. 1307--1314.

\bibitem{DuToit2012}
N.~E. {Du Toit} and J.~W. {Burdick}, ``Robot motion planning in dynamic,
  uncertain environments,'' \emph{IEEE Transactions on Robotics}, vol.~28,
  no.~1, pp. 101--115, Feb 2012.

\bibitem{Lee2017}
W.~Lee and D.~Kim, ``Autonomous shepherding behaviors of multiple target
  steering robots,'' \emph{Sensors}, vol.~17, no.~12, 2017.

\bibitem{Tsunoda2017}
Y.~{Tsunoda}, Y.~{Sueoka}, and K.~{Osuka}, ``On statistical analysis for
  shepherd guidance system,'' in \emph{2017 IEEE International Conference on
  Robotics and Biomimetics (ROBIO)}, Dec 2017, pp. 1246--1251.

\bibitem{Fujioka2016}
K.~Fujioka and S.~Hayashi, ``Effective shepherding behaviours using multi-agent
  systems,'' in \emph{IEEE Region 10 Conference (TENCON)}, Singapore, Nov 2016,
  pp. 3179--3182.

\bibitem{Fujioka2018}
K.~Fujioka, ``Effective herding in shepherding problem in {V}-formation
  control,'' \emph{Transactions of the Institute of Systems, Control and
  Information Engineers}, vol.~31, no.~1, pp. 21--27, 2018.

\bibitem{Reynolds1987}
C.~W. Reynolds, ``Flocks, herds and schools: A distributed behavioral model,''
  \emph{{SIGGRAPH} Comput. Graph.}, vol.~21, no.~4, pp. 25--34, Aug 1987.

\bibitem{Reynolds1999}
------, ``Steering behaviors for autonomous characters,'' in \emph{Game
  Developers Conference}, vol. 1999, 1999, pp. 763--782.

\bibitem{Bayazit2002}
O.~{Burchan Bayazit}, {Jyh-Ming Lien}, and N.~M. {Amato}, ``Roadmap-based
  flocking for complex environments,'' in \emph{10th Pacific Conference on
  Computer Graphics and Applications, 2002. Proceedings.}, Oct 2002, pp.
  104--113.

\bibitem{Bayazit2004}
O.~B. Bayazit, J.-M. Lien, and N.~M. Amato, \emph{Better Group Behaviors Using
  Rule-Based Roadmaps}.\hskip 1em plus 0.5em minus 0.4em\relax Berlin,
  Heidelberg: Springer Berlin Heidelberg, 2002, pp. 95--111.

\bibitem{Lien2005}
J.-M. Lien, S.~Rodr{\'i}guez, J.-P. Malric, and N.~M. Amato, ``Shepherding
  behaviors with multiple shepherds,'' in \emph{Proceedings of the 2005 IEEE
  International Conference on Robotics and Automation}, 2005, pp. 3402--3407.

\bibitem{Harrison2010}
J.~F. Harrison, C.~Vo, and J.-M. Lien, ``Scalable and robust shepherding via
  deformable shapes,'' in \emph{Motion in Games}, R.~Boulic, Y.~Chrysanthou,
  and T.~Komura, Eds.\hskip 1em plus 0.5em minus 0.4em\relax Berlin,
  Heidelberg: Springer Berlin Heidelberg, 2010, pp. 218--229.

\bibitem{Razali2010}
S.~{Razali}, Q.~{Meng}, and S.~{Yang}, ``A refined immune systems inspired
  model for multi-robot shepherding,'' in \emph{Second World Congress on Nature
  and Biologically Inspired Computing (NaBIC)}, Dec 2010, pp. 473--478.

\bibitem{Bacon2011}
M.~Bacon and N.~Olgac, ``Swarm herding using a region holding sliding mode
  controller,'' \emph{Journal of Vibration and Control}, vol.~18, no.~7, pp.
  1056--1066, 2011.

\bibitem{Pierson2018}
A.~{Pierson} and M.~{Schwager}, ``Controlling noncooperative herds with robotic
  herders,'' \emph{IEEE Transactions on Robotics}, vol.~34, no.~2, pp.
  517--525, Apr 2018.

\bibitem{Tsunoda2018}
Y.~Tsunoda, Y.~Sueoka, Y.~Sato, and K.~Osuka, ``Analysis of local-camera-based
  shepherding navigation,'' \emph{Advanced Robotics}, vol.~32, no.~23, pp.
  1217--1228, 2018.

\bibitem{Cimler2016}
R.~Cimler, O.~Dole{\v{z}}al, J.~K{\"u}hnov{\'a}, and J.~Pavl{\'i}k, ``Herding
  algorithm in a large scale multi-agent simulation,'' in \emph{Agent and
  Multi-Agent Systems: Technology and Applications}, G.~Jezic, Y.-H.~J.
  Chen-Burger, R.~J. Howlett, and L.~C. Jain, Eds.\hskip 1em plus 0.5em minus
  0.4em\relax Cham: Springer International Publishing, 2016, pp. 83--94.

\bibitem{Razali2013}
S.~{Razali}, N.~F. {Shamsudin}, M.~{Osman}, Q.~{Meng}, and S.~{Yang}, ``Flock
  identification using connected components labeling for multi-robot
  shepherding,'' in \emph{International Conference on Soft Computing and
  Pattern Recognition (SoCPaR)}, Dec 2013, pp. 298--303.

\bibitem{Long2019}
N.~Long, D.~Sgarioto, M.~Garratt, K.~Sammut, and H.~Abbass, ``Multi-vessel sea
  state estimation utilising swarm shepherding.'' in \emph{Proceedings of 2019
  Pacific International Maritime Conference}, Oct 2019.

\bibitem{Long20192}
N.~Long, M.~Garratt, D.~Sgarioto, K.~Sammut, and H.~Abbass, \emph{Shepherding
  Autonomous Goal-Focused Swarms in Unknown Environments Using Space-Filling
  Paths}.\hskip 1em plus 0.5em minus 0.4em\relax Springer International
  Publishing, Accepted.

\bibitem{ATSB2017}
\BIBentryALTinterwordspacing
{Australian Transport Safety Bureau}, ``Aviation occurrence statistics,'' 2017.
  [Online]. Available:
  \url{https://www.atsb.gov.au/media/5773880/ar-2017-104$\_$final.pdf}
\BIBentrySTDinterwordspacing

\bibitem{FAA2019}
\BIBentryALTinterwordspacing
{Federal Aviation Safety Authority}, ``{FAA Wildlife Strike Database},'' 2019.
  [Online]. Available: \url{https://wildlife.faa.gov/home}
\BIBentrySTDinterwordspacing

\bibitem{Nalepka2017}
P.~Nalepka, R.~W. Kallen, A.~Chemero, E.~Saltzman, and M.~J. Richardson, ``Herd
  those sheep: Emergent multiagent coordination and behavioral-mode
  switching,'' \emph{Psychological Science}, vol.~28, no.~5, pp. 630--650,
  2017.

\bibitem{Nalepka2016}
P.~Nalepka, M.~Lamb, R.~W. Kallen, K.~Shockley, A.~Chemero, and M.~J.
  Richardson, ``A bio-inspired artificial agent to complete a herding task with
  novices,'' in \emph{Conference on Artificial Life 2016}, no.~28, 2016, pp.
  656--663.

\bibitem{Nalepka2018}
P.~Nalepka, R.~W. Kallen, M.~Lamb, and M.~J. Richardson, ``Emergence of
  efficient, coordinated solutions despite differences in agent ability during
  human-machine interaction: Demonstration using a multiagent ``shepherding''
  task,'' in \emph{Proceedings of the 18th International Conference on
  Intelligent Virtual Agents}.\hskip 1em plus 0.5em minus 0.4em\relax New York,
  USA: ACM, 2018, pp. 337--338.

\bibitem{Nalepka2019}
P.~Nalepka, R.~W. Kallen, A.~Chemero, E.~Saltzman, and M.~J. Richardson,
  ``Practical applications of multiagent shepherding for human-machine
  interaction,'' in \emph{Advances in Practical Applications of Survivable
  Agents and Multi-Agent Systems: The PAAMS Collection}, Y.~Demazeau,
  E.~Matson, J.~M. Corchado, and F.~De~la Prieta, Eds.\hskip 1em plus 0.5em
  minus 0.4em\relax Cham: Springer International Publishing, 2019, pp.
  168--179.

\bibitem{Tuyls2016}
K.~Tuyls, S.~Alers, E.~Cucco, D.~Claes, and D.~Bloembergen, ``A
  telepresence-robot approach for efficient coordination of swarms,''
  \emph{Conference on Artificial Life}, no.~28, pp. 666--673, 2016.

\bibitem{Wade2019CyberShepherd}
H.~Wade and H.~A. Abbass, ``Cyber-shepherd: A smartphone-based game for human
  and autonomous swarm control,'' in \emph{IEEE Systems, Man, Cybernetics
  Conference}.\hskip 1em plus 0.5em minus 0.4em\relax Bari, Italy: IEEE, 2019.

\bibitem{alex2019machine}
A.~Gee and H.~A. Abbass, ``Transparent machine education of neural networks for
  swarm shepherding using curriculum design,'' in \emph{International Joint
  Conference on Neural Networks}.\hskip 1em plus 0.5em minus 0.4em\relax
  Budapest, Hungary: IEEE, 2019.

\bibitem{clayton2019machine}
N.~R. Clayton and H.~A. Abbass, ``Machine teaching in hierarchical genetic
  reinforcement learning: Curriculum design of reward functions for swarm
  shepherding,'' in \emph{IEEE Congress on Evolutionary Computation}.\hskip 1em
  plus 0.5em minus 0.4em\relax Wellington, New Zealand: IEEE, 2019.

\bibitem{Skrba2006}
L.~Skrba, S.~Dobbyn, R.~McDonnell, and C.~O'Sullivan, ``Animating dolly:
  Real-time herding and rendering of sheep,'' in \emph{Seventh Irish Workshop
  on Computer Graphics (2006)}, Jan 2006.

\bibitem{Nguyen2018QLearning}
T.~{Nguyen}, H.~{Nguyen}, E.~{Debie}, K.~{Kasmarik}, M.~{Garratt}, and
  H.~{Abbass}, ``Swarm {Q}-leaming with knowledge sharing within environments
  for formation control,'' in \emph{International Joint Conference on Neural
  Networks (IJCNN)}, Jul 2018, pp. 1--8.

\bibitem{Elman1993}
J.~L. Elman, ``Learning and development in neural networks: the importance of
  starting small,'' \emph{Cognition}, vol.~48, 1993.

\bibitem{Abbass2019}
H.~A. Abbass, S.~Elsawah, E.~Petraki, and R.~Hunjet, ``Machine education:
  Designing semantically ordered and ontologically guided modular neural
  networks,'' in \emph{IEEE Symposium Series on Computational Intelligence},
  Xiamen, China, Dec 2019.

\bibitem{Nguyen2018Apprenticeship}
H.~Nguyen, M.~Garratt, and H.~Abbass, ``Apprenticeship bootstrapping,'' in
  \emph{International Joint Conference on Neural Networks (IJCNN)}.\hskip 1em
  plus 0.5em minus 0.4em\relax IEEE, 2018, pp. 1--8.

\bibitem{Nguyen2018Apprenticeship2}
H.~Nguyen, M.~Garratt, L.~T. Bui, and H.~Abbass, ``Apprenticeship
  bootstrapping: Inverse reinforcement learning in a multi-skill {UAV-UGV}
  coordination task,'' in \emph{Proceedings of the 17th International
  Conference on Autonomous Agents and MultiAgent Systems}.\hskip 1em plus 0.5em
  minus 0.4em\relax International Foundation for Autonomous Agents and
  Multiagent Systems, 2018, pp. 2204--2206.

\bibitem{Nguyen2019ICONIP}
T.~{Nguyen}, H.~{Nguyen}, E.~{Debie}, K.~{Kasmarik}, M.~{Garratt}, and
  H.~{Abbass}, ``A deep hierarchical reinforcement learner for aerial
  shepherding of ground swarms,'' in \emph{International Conference on Neural
  Information Processing}, Dec 2019.

\bibitem{Singh2019}
H.~Singh, B.~Campbell, S.~Elsayed, A.~Perry, R.~Hunjet, and H.~Abbass,
  ``Modulation of force vectors for effective shepherding of a swarm: A
  bi-objective approach,'' in \emph{IEEE Congress on Evolutionary Computation
  (CEC)}.\hskip 1em plus 0.5em minus 0.4em\relax IEEE, 2019, pp. 2941--2948.

\bibitem{Abbass2018}
H.~A. Abbass, J.~Harvey, and K.~Yaxley, ``Lifelong testing of smart autonomous
  systems by shepherding a swarm of watchdog artificial intelligence agents,''
  \emph{Computing Research Repository}, vol. abs/1812.08960, 2018.

\bibitem{Celikkanat2010}
H.~{\c{C}}elikkanat and E.~{\c{S}}ahin, ``Steering self-organized robot flocks
  through externally guided individuals,'' \emph{Neural Computing and
  Applications}, vol.~19, no.~6, pp. 849--865, Sep 2010.

\bibitem{tang2017networking}
J.~Tang, G.~Leu, and H.~A. Abbass, ``Networking the boids is more robust
  against adversarial learning,'' \emph{IEEE Transactions on Network Science
  and Engineering}, vol.~5, no.~2, pp. 141--155, 2018.

\end{thebibliography}

\begin{IEEEbiography}[{\includegraphics[width=1in,height=1.25in,clip,keepaspectratio]{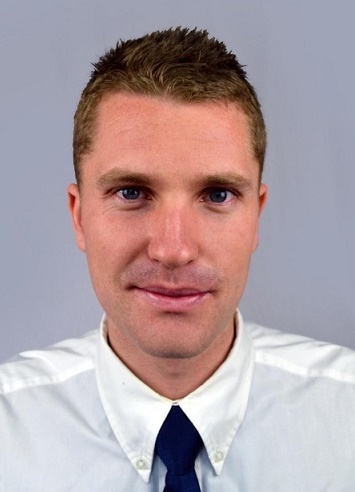}}]{Nathan K. Long} is currently undertaking a Masters by Research in Computer Science at the University of New South Wales on the topic of sea state estimation using a swarm of uninhabited surface vessels, guided by a shepherd. His masters is being sponsored by the Defence Science and Technology Group, Maritime Division, in the form of a Cadetship. Nathan graduated from a Bachelor of Engineering (Aerospace), with a First Class Honours from RMIT University in 2018. He now sits on the American Institute of Aeronautics and Astronautics' (AIAA) Sydney Section Committee, and is a student member of IEEE. 
\end{IEEEbiography}

\begin{IEEEbiography}[{{{\includegraphics[clip,width=1in,height=1in,keepaspectratio]{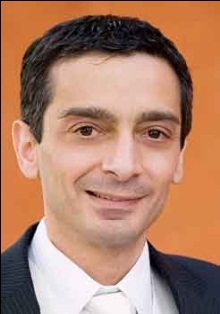}}}}]{Karl Sammut}
completed his PhD at The University of Nottingham (U.K) in 1992 and was employed between 1992 and 1995 as a Postdoctoral Fellow with the Politecnico di Milano (Italy), and with Loughborough University (UK). He commenced his appointment at Flinders University in 1995 and is currently a Professor in the College of Science and Engineering. He serves as the Director of the Centre for Maritime Engineering at Flinders University and the Theme Leader for the Maritime Autonomy Group. He also holds a part-time position with DST Group Maritime Division. His areas of research specialization are concerned with navigation, optimal guidance and control systems, and mission planning systems for autonomous marine surface and underwater vehicles.  
\end{IEEEbiography}

\begin{IEEEbiography}[{{{\includegraphics[clip,width=1in,height=1in,keepaspectratio]{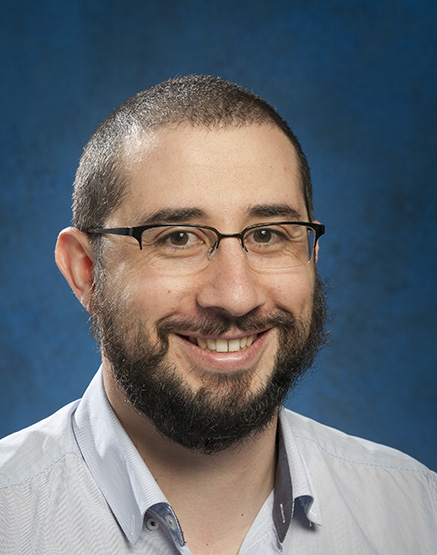}}}}]{Daniel Sgarioto}
received a PhD in aerospace engineering from RMIT University in 2006. From 2006 to 2011 he was a scientist at the Defence Technology Agency in New Zealand, where his research focused on the modeling and control of autonomous underwater vehicles. In 2012 he joined the Defence Science and Technology Group where his current research interests include multi-vessel interactions, seakeeping performance and operability of naval vessels. 
\end{IEEEbiography}

\begin{IEEEbiography}[{{{\includegraphics[clip,width=1in,height=1in,keepaspectratio]{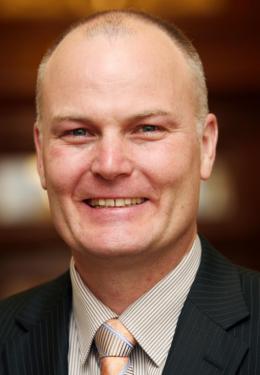}}}}]{Matthew Garratt}
is the Deputy Head of School (Research) for the school of Engineering and IT and is based in the Canberra campus of UNSW Australia. His main research areas focus on sensing, guidance and control for autonomous systems. He is a member of the Trusted Autonomy Group in the School of Engineering and IT, as well as being a senior lifetime member of the American Institute of Aeronautics and Astronautics (AIAA), senior member of the IEEE, member of the American Helicopter Society (AHS), and member of the Australian Association for Unmanned Systems (AAUS).
\end{IEEEbiography}

\begin{IEEEbiography}[{{{\includegraphics[clip,width=1in,height=1in,keepaspectratio]{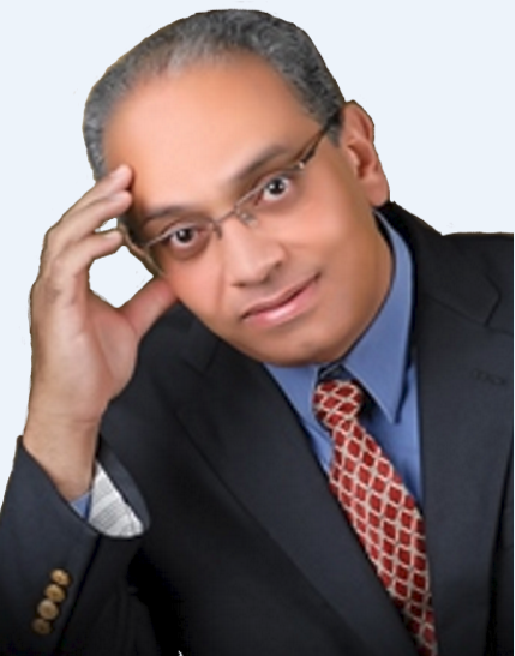}}}}]{Hussein A. Abbass}
(F\textquoteright 20) is a Professor at the University of New South Wales, Canberra, Australia. He is a fellow of the IEEE, a fellow of the
Australian Computer Society (FACS), a fellow of the Operational Research Society (FORS,UK); and a
fellow of the Institute Management and Leaders of Australia (FIML). He was the Vice-president for Technical
Activities (2016-2019) for the IEEE Computational Intelligence Society, and the National President for the Australian Operations Research Society (2016-2019). He is the Founding Editor-in-Chief of the IEEE Transactions on Artificial Intelligence. His current research contributes to trusted human-swarm teaming with an aim to design next generation trusted and distributed artificial intelligence systems that seamlessly integrate humans and machines.

\end{IEEEbiography}

\vfill 

\end{document}